%% file: neurips_2024.tex
\documentclass{article}

\PassOptionsToPackage{numbers, compress}{natbib}

\usepackage[preprint]{neurips_2024}

\usepackage[utf8]{inputenc} %
\usepackage[T1]{fontenc}    %
\usepackage{hyperref}       %
\usepackage{url}            %
\usepackage{booktabs}       %
\usepackage{amsfonts}       %
\usepackage{mathtools} %
\usepackage{nicefrac}       %
\usepackage{microtype}      %
\usepackage{xcolor}         %
\usepackage{listings}

\usepackage{subcaption}
\usepackage{graphicx}
\usepackage{graphbox}
\usepackage{amssymb} %
\usepackage{pifont}  %
\usepackage[inline]{enumitem}
\usepackage{wrapfig}
\usepackage{booktabs} %
\usepackage{array}    %
\usepackage{float} %

\usepackage{color}

\definecolor{codegreen}{rgb}{0,0.6,0}
\definecolor{codegray}{rgb}{0.5,0.5,0.5}
\definecolor{codepurple}{rgb}{0.58,0,0.82}
\definecolor{backcolour}{rgb}{0.95,0.95,0.92}

\lstdefinestyle{mystyle}{
    backgroundcolor=\color{backcolour},   
    commentstyle=\color{codegreen},
    keywordstyle=\color{magenta},
    numberstyle=\tiny\color{codegray},
    stringstyle=\color{codepurple},
    basicstyle=\tiny\ttfamily,
    breakatwhitespace=false,         
    breaklines=true,                 
    captionpos=b,                    
    keepspaces=true,                 
    numbers=left,                    
    numbersep=5pt,                  
    showspaces=false,                
    showstringspaces=false,
    showtabs=false,                  
    tabsize=2
}

\definecolor{darkred}{rgb}{0.6, 0.0, 0.0}  %
\definecolor{Blue}{rgb}{0.2,0.2,0.6}
\hypersetup{
  colorlinks=true,
  citecolor=Blue,
  linkcolor=Blue,
  urlcolor=Blue,
}%
\usepackage{tikz} %
\usetikzlibrary{fit,positioning,arrows,automata}

\usepackage{cleveref}
\crefname{figure}{Fig.}{Figs.}
\Crefname{figure}{Fig.}{Figs.}
\crefname{section}{Sec.}{Sec.}
\Crefname{section}{Sec.}{Sec.}
\crefname{table}{Table}{Table}
\Crefname{table}{Table}{Table}
\crefname{listing}{Listing}{Listing}

\newcommand{\cmark}{\textcolor{green}{\ding{51}}} %
\newcommand{\xmark}{\textcolor{red}{\ding{55}}}   %
\newcommand{\ours}{\texttt{RefDrop}}

\newcommand{\rma}{\texttt{RFG}}
\newcommand{\RMA}{\text{Reference Feature Guidance}}

\newcommand{\mR}{{\mathbb R}}
\def\<{{\langle}}
\def\>{{\rangle}}

\title{RefDrop: Controllable Consistency in Image or Video Generation via Reference Feature Guidance}

\author{%
  Jiaojiao Fan
    \\
  Georgia Tech\\
  \texttt{sbyebss@gmail.com} \\
  \And
  Haotian Xue \\
  Georgia Tech \\
  \texttt{htxue.ai@gatech.edu} \\
  \AND
  Qinsheng Zhang\thanks{Work done while the author was in Georgia Tech.}  \\
  NVIDIA \\
  \texttt{qsh.zh27@gmail.com} \\
  \And
  Yongxin Chen \\
  Georgia Tech \\
  \texttt{yongchen@gatech.edu} \\
}

\begin{document}

\maketitle

\input{main.tex}

\input{supple.tex}

\end{document}

%% file: main.tex
\begin{abstract}
There is a rapidly growing interest in controlling consistency across multiple generated images using diffusion models. Among various methods, recent works have found that simply manipulating attention modules by concatenating features from multiple reference images provides an efficient approach to enhancing consistency without fine-tuning. Despite its popularity and success, few studies have elucidated the underlying mechanisms that contribute to its effectiveness. In this work, we reveal that the popular approach is a linear interpolation of image self-attention and cross-attention between synthesized content and reference features, with a constant rank-1 coefficient. Motivated by this observation, we find that a rank-1 coefficient is not necessary and simplifies the controllable generation mechanism. The resulting algorithm, which we coin as \ours, allows users to control the influence of reference context in a direct and precise manner. Besides further enhancing consistency in single-subject image generation, our method also enables more interesting applications, such as the consistent generation of multiple subjects, suppressing specific features to encourage more diverse content, and high-quality personalized video generation by boosting temporal consistency. Even compared with state-of-the-art image-prompt-based generators, such as IP-Adapter, \ours ~is competitive in terms of controllability and quality while avoiding the need to train a separate image encoder for feature injection from reference images, making it a versatile plug-and-play solution for any image or video diffusion model. Our project webpage is \url{https://sbyebss.github.io/refdrop/}.
\end{abstract}

\begin{figure}[h]
  \centering
    \makebox[\linewidth]{\includegraphics[width=1\linewidth]{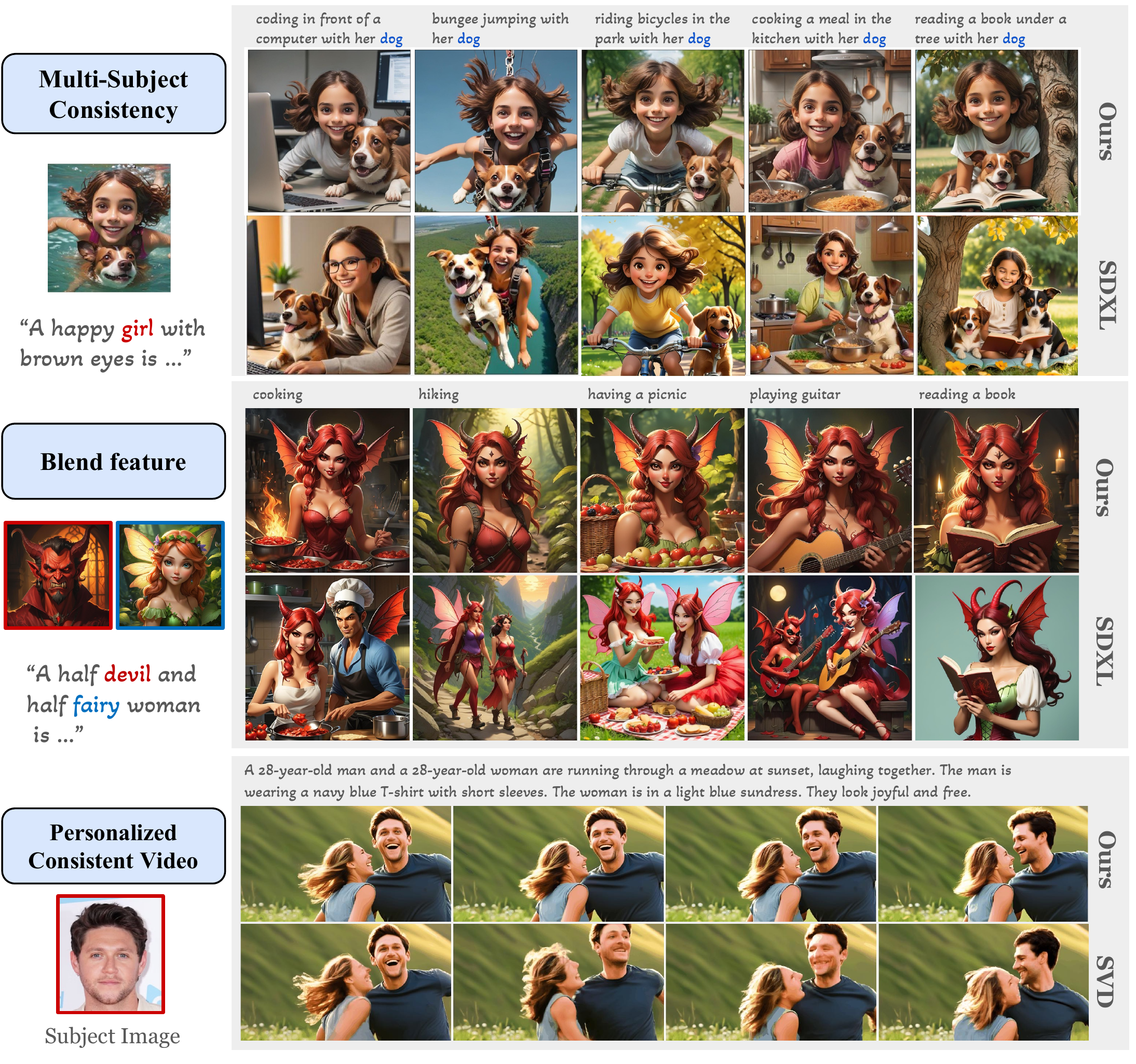}}
  \caption{
\ours ~achieves controllable consistency in visual content synthesis for free.
\ours ~exihibits great flexibility in
(Upper) multi-subject consistency generation given one reference image,
(Middle) blending different characters from multiple images seamlessly,
(Buttom) enhancing temporal consistency for personalized video generation.
   \ours ~is short for "reference drop". We named our method \ours ~to metaphorically represent the process by which a drop of colored water influences a larger body of clear water.
 }
\end{figure}

\section{Introduction}

Large-scale diffusion models have demonstrated remarkable capabilities in aiding content creation for artists~\citep{podell2023sdxl,sora,balaji2022ediff}. Numerous text-to-image models are expediting content production in various domains, including advertising and art studios. Similarly, video generation models have shown significant advancements recently~\citep{girdhar2023emu,chen2023videocrafter1,guo2023animatediff,Hu2023-wr, makeavideo, blattmann2023align, bar2024lumiere, vdm}. However, enhancing these models to better support artistic creativity requires improved controllability, particularly in content consistency. This paper explores consistency from two perspectives: 1) controlling subject consistency across multiple images, and 2) maintaining subject consistency across multiple frames within a video. 

We name a few tasks where the controllable consistency is crucial in AI content generation.
In image generation for storytelling~\citep{rahman2023make,maharana2022storydall, pan2024synthesizing, jeong2023zero} or advertising, content creators often strive to produce consistent characters, a task that proves challenging with foundational generative models~\citep{tewel2024training}. Personalization approaches based on fine-tuning~\citep{ruiz2023dreambooth} require a minimum of 5 to 10 images to achieve satisfactory quality, and encoder-based methods~\citep{wei2023elite,wang2024instantid, pan2023kosmos} demand weeks of training 
with millions of images
for a single diffusion model and lack transferability to other foundational models. 
On the other hand, diverse image generation is less addressed but persistently challenging. In this scenario, it is desired to \textit{decrease} the consistency among image generations. For example,
artists can sometimes seek to enhance diversity and avoid clichés, such as the stereotypical depiction of Barbie girls with curly blonde hair. 
For video generation, another challenging task is
maintaining temporal consistency in video generation, yet most existing solutions are confined to video editing tasks~\citep{ku2024anyv2v}, demanding high-quality input videos. 

These emerging tasks motivate us to develop \ours, a \textbf{
training-free}, \textbf{plug-and-play} method designed to 
provide flexible control over the consistency in image and video generation.
Specifically, we modify the self-attention mechanism in the diffusion model UNet~\citep{Ronneberger2015-iw} architecture and introduce a coefficient to modulate the influence of a reference image on the generation process. Our contributions are outlined as follows:

1. We conduct a detailed analysis of popular consistency generation methods based on concatenated attention, revealing that their consistency is actually contributed by extra guidance applied implicitly.
2. Inspired by this finding, we propose \RMA~(\rma), a natural extension that explicitly controls the guidance from reference context in a precise and direct manner. Building upon~\rma, we introduce \ours, a flexible and efficient approach to controlling consistency without the need for network fine-tuning or optimization.
3. Besides improvements in character consistency using a single reference image, \ours ~enables more creative applications with controllable consistency, including (i) seamless integration of distinct features into a single cohesive image (ii) suppressing specific features by negatively decreasing the consistency influenced by the reference context, thereby enhancing diversity in layout, accessories, and image style; (iii) high-quality personalized video generation by boosting temporal consistency,
and minimizing facial distortions. 
4. We conduct comprehensive experiments and demonstrate that \ours ~achieves a good balance between flexibility and effectiveness while being lightweight compared to existing works.

\section{Related work}
Among the works most similar to ours are IP-Adapter~\citep{ye2023ipa} and concatenated attention~\citep{wu2023tune}. Our approach is closely related to IP-Adapter, as both methods utilize the sum of two decoupled attention outputs. However, while IP-Adapter modifies cross-attention and requires separate training of an image encoder to embed the reference image, we integrate the reference image directly into the self-attention layer without needing additional training. Furthermore, our reference images are \textit{generated} by the same model, in contrast to IP-Adapter's reliance on externally sourced image. Both techniques permit the use of negative or positive coefficients for the reference image, but IP-Adapter may compromise text alignment~\cite{tewel2024training} due to its reference image being intertwined with the text prompt during cross-attention. Additionally, the IP-Adapter requires separate training for different versions of the diffusion model, such as SD2.1 and SDXL. In contrast, \ours ~is a simple plug-and-play.

Concatenated attention, first introduced in video generation literature by \citet{wu2023tune} as spatio-temporal attention, injects temporal information into a T2I model. It has since been widely adopted for feature injection across various applications \citep{luo2024diffusion, chang2023magicdance, huang2024parts, tewel2024training} in content generation and video editing. This concept has evolved into Cross-Frame Attention~\citep{khachatryan2023text2video}, another prevalent technique used to inflate T2I models~\citep{zhang2024videoelevator} for video generation. We will demonstrate later that our framework can replicate these two types of attention as special cases.

\vspace{-0.2cm}
\paragraph{Consistency in Image Generation}
ConsiStory \citep{tewel2024training} and StoryDiffusion \citep{zhou2024storydiffusion} are closely related to our work. They are training-free methods that employs concatenated attention to enhance consistency in generation. Our \rma ~framework is {\em orthogonal} to the techniques other than concatenated attention in those works, such as subject masking and attention dropout.
\citet{avrahami2023chosen} explores a fine-tuning-based method aimed at recovering tightly clustered images. 
Other approaches, such as those by \citep{jeong2023zero,feng2023improved,liu2023intelligent}, predominantly utilize a personalization process \citep{ruiz2023dreambooth,gal2022image,kumari2023multi,tewel2023key} requiring multiple input images for training.
Finally, several encoder-based methods \citep{wei2023elite,wang2024instantid,li2024blip,ruiz2023hyperdreambooth,xiao2023fastcomposer,kim2024instantfamily,li2023photomaker} do not require additional training for new subjects. However, these methods necessitate days or weeks of initial training for the encoder and face limitations in adaptability to different versions of foundational generative models.

\paragraph{Temporal-consistency in video generation}
Concatenated attention~\citep{wu2023tune,ren2024consisti2v} and Cross-Frame Attention~\citep{khachatryan2023text2video,Chen2024-ag} are popular techniques used to inflate T2I models for video generation. \citet{wu2023freeinit, ren2024consisti2v} mitigate video flickering by applying a low-pass filter to noisy latent images, effectively removing disruptive high-frequency content. Many other methods are tailored for video editing tasks, and they either extract features from high-quality input videos to enhance the current generation \citep{ku2024anyv2v, zhang2024fastvideoedit, yang2024fresco, yang2023rerender} or use them as references during editing \citep{geyer2023tokenflow, cong2023flatten}. \ours ~improves temporal consistency directly within video generation, obviating the need for an input video.

\section{Method}
\label{headings}
We first introduce how existing works achieve consistency generation by leveraging the concatenation of reference features in the self-attention block. Then we reformulate the concatenation as a linear interpolation of self-attention on synthesized content and cross-attention between generated and reference content with a constant rank-1 coefficient. We highlight that this specific coefficient is not a necessity, while linear interpolation is critical to minimizing the training-inference gap. Building upon these observations, we propose \RMA~(\rma), an extension of concatenation attention that allows for flexible feature interpolation and extrapolation in attention modules. Based on~\rma, we introduce~\ours, a versatile method for controllable consistency generation across various applications.

\begin{figure}[t!]
  \centering
  \vspace{-0.2cm}
    \makebox[\linewidth]{\includegraphics[width=1\linewidth]{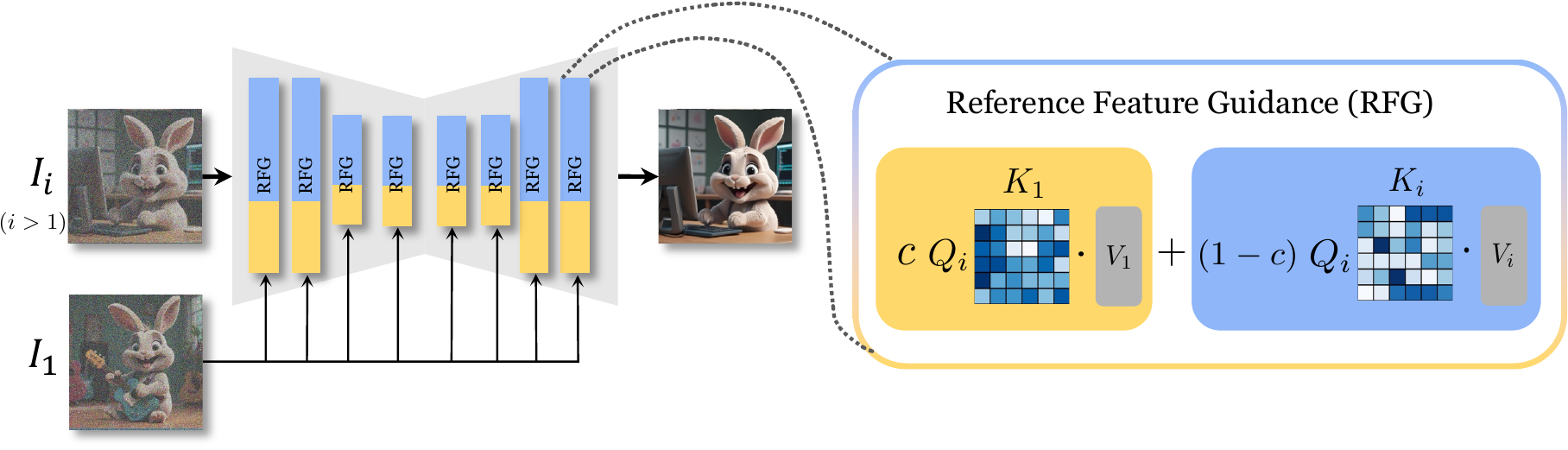}}
  \caption{During each diffusion denoising step, we facilitate the injection of features from a \textit{generated} reference image $I_1$ into the generation process of other images through \rma. The \rma ~layer produces a linear combination of the attention outputs from both the standard and referenced routes. A negative coefficient $c$ encourages divergence of $I_i$ from $I_1$, while a positive coefficient fosters consistency.
  }\label{fig:rma}
\end{figure}

\subsection{Background}
Self-attention in diffusion model networks operates by applying the attention mechanism~\citep{Vaswani2017-hk} on synthesized latent features.
A self-attention layer processes latent representations \(X\) by passing them through linear projection layers to produce queries $Q = X W_Q$, keys $K = X W_K$, and values $V = X W_V$, which then undergo the attention operation as follows:
\begin{align}
     X' = \text{Attention} (Q, K, V) = \text{Softmax}\left(\frac{QK^\top}{\sqrt{d}}\right) V, \label{eq:self-attn}
\end{align}
where $X'$ is the output of self-attention operation, and $d$ is the feature dimension of projection matrices $W_Q, W_K,W_V.$
Previous consistency generation~\citep{tewel2024training,zhou2024storydiffusion} is based on concatenated attention via a simple batch image generation, where the first sample in the batch serves as a reference for the $i$-th sample generation.
We denote the latent feature for $i$-th sample as $X_i$. Instead of solely depending on its own content, concatenated attention suggests
\begin{equation}\label{eq:concat}
X_{\texttt{CAT}}' =  
\text{Attention}\left({Q_i, [K_1; K_i], [V_1; V_i]}\right),
\end{equation}
where \(Q_i = X_i W_Q\), \(K_i = X_i W_K\), and \(V_i = X_i W_V\).

\subsection{Reference feature guidance}\label{sec:rma}

To illustrate why concatenated attention can help boost consistency between generated samples with reference samples, we can reformulate~\cref{eq:concat} as the following~(Proof in~\cref{sec:relation_concat})
\begin{align}\label{eq:concat_rewrite}
X_{\texttt{CAT}}' &=  
C \odot \text{Attention}\left({Q_i, K_1, V_1}\right) + (\mathbf{1}-C) \odot \text{Attention}\left({Q_i, K_i, V_i}\right)
\end{align}
where \(C\) is a rank-1 matrix of the same size as the attention output, \(\odot\) is the point-wise multiplication and \(\mathbf{1}\) is an all-ones matrix.

\Cref{eq:concat_rewrite} depicts that the concatenated attention is a linear interpolation between the output $X'$ without concatenated attention in~\cref{eq:self-attn} and cross-attention between the $i$-th image $X_i$ and the reference image $X_1$, while the coefficient matrix $C$ is determined by the synthesized content $X_i$ and the reference content $X_1$. Before we further improve concatenated attention, we first discuss two related questions for~\cref{eq:concat_rewrite}.
\textbf{Is linear interpolation a necessity?} 
It may be tempting to highlight the role of the second cross-attention term naively while keeping the weights for the first term unchanged, such as $\text{Attention}\left(Q_i, K_1, V_1\right) + \text{Attention}\left(Q_i, K_i, V_i\right)$. However, we find that naively breaking the linear interpolation disrupts image generation. In fact, we can interpret concatenated attention in \cref{eq:concat_rewrite} as applying extra guidance on the original self-attention output 
\begin{align}\label{eq:guidance}
X_{\texttt{CAT}}'
&= \text{Attention}\left({Q_i, K_i, V_i}\right) + C \odot ( \text{Attention}\left({Q_i, K_1, V_1}\right) - \text{Attention}\left({Q_i, K_i, V_i}\right))
\end{align}
which resembles the form of guidance used in diffusion literature~\citep{song2023loss,ho2022video}, such as classifier-free guidance~\citep{ho2022classifier}. 
Notably, the linear interpolation helps keep the attention output $X_{\texttt{CAT}}'$ norm close to self attention output $X'$; otherwise, arbitrary weights would pose a training and inference discrepancy and degrade the generation quality.
However, different from various guidance methods used in the diffusion literature, the guidance weights in \cref{eq:guidance} are constants determined by latent features $X_i$ and reference context $X_1$ and have no user control. 
Therefore we question 
\textbf{Is constant $C$ matrix coefficient is a necessity?} 
As an attempt to bypass the rigid form of concatenated attention, we propose a simple and flexible approach named \textit{Reference feature guidance}~(\rma) (see \cref{fig:rma})
\begin{equation}
X_{\rma}' = c \cdot \text{Attention}\left({Q_i, K_1, V_1}\right) + (1-c) \cdot \text{Attention}\left({Q_i, K_i, V_i}\right),
\label{eq:rma}
\end{equation}
where \(c\) is a scalar coefficient that controls the strength of the reference image influence.

\begin{wrapfigure}[16]{r}{0.6\textwidth}
\vspace{-0.5cm}
\centering
    \includegraphics[width=1\linewidth , align = c]{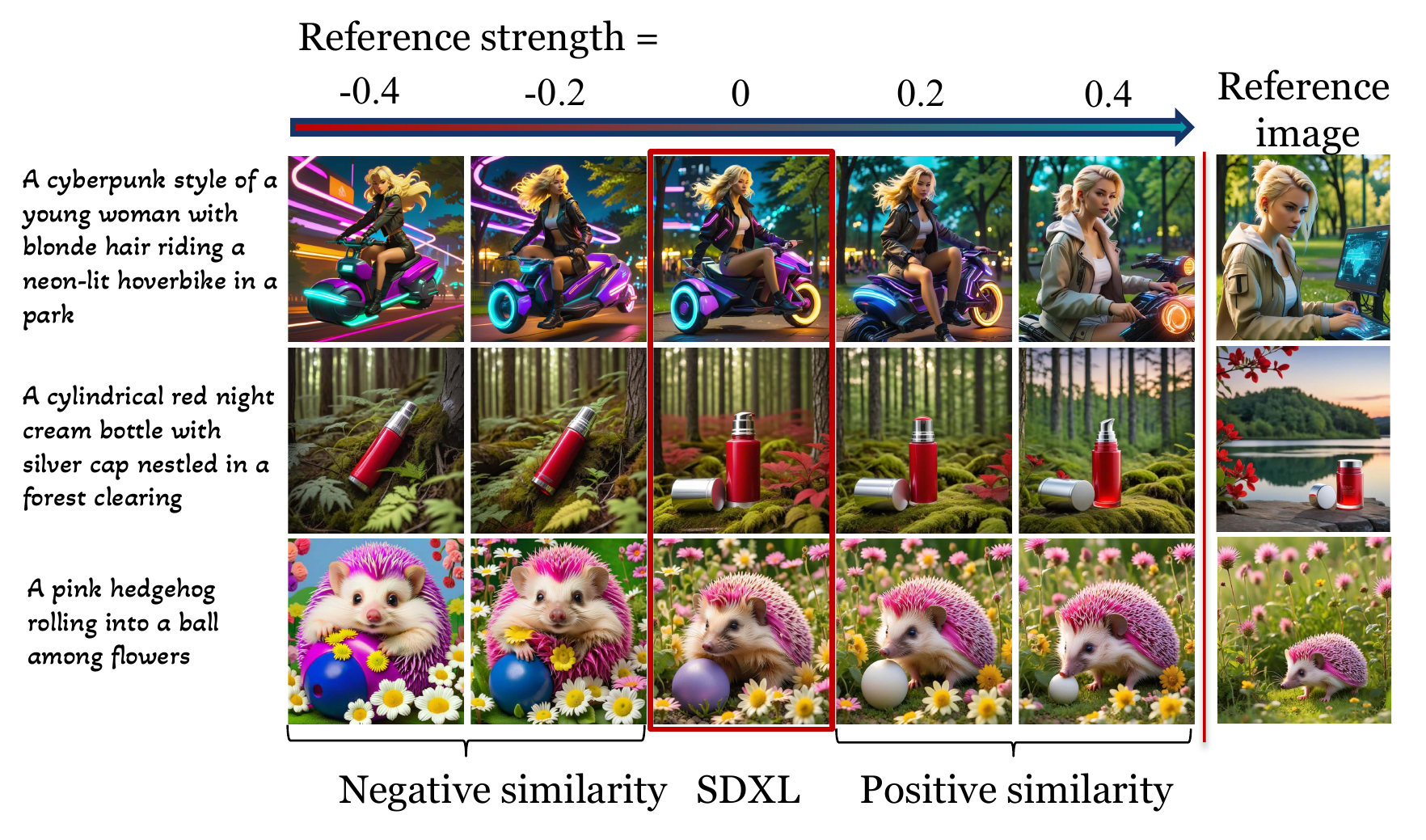}   
    \caption{We allow flexible control over the reference effect through a reference strength coefficient. 
    }  \label{fig:coeff}
\end{wrapfigure} 
While most previous methods for consistent generation, including feature combination~\citep{kim2024instantfamily,zhou2024storydiffusion} and injection~\citep{tewel2024training}, employ concatenated attention~\eqref{eq:concat}, our \rma~offers several advantages. First, it grants users greater control over the extent of influence from the reference image, as illustrated in \cref{fig:coeff}.
Second, this flexibility proves especially beneficial in a novel application: blending features from multiple reference images. Our method allows users to selectively determine the influence of each reference image. We have observed that the most harmonious blending often results from varying the strength of each reference, rather than maintaining equal strength across all images. Third, by enabling negative coefficients, we find that our method can simulate a concept suppression effect, meaning it generates images that are dissimilar to a reference image. Finally, it allows for the injection of reference image features into video generation slightly to reduce flickering, while the concatenated attention keeps the video completely static.

Therefore, we introduce~\ours, a training-free approach to flexibly control consistency generation, which replaces the self-attention blocks in the diffusion model with~\rma.
For Video Diffusion Models (VDM)~\citep{blattmann2023stable, ge2023preserve, he2022latent}, we modify \textit{every} \textit{spatial} self-attention layers to bolster temporal consistency.

\begin{wraptable}[11]{r}{0.5\textwidth} 
  \vspace{-1.3cm}
  \caption{Comparison of Controllable Consistent Image Generation Methods. `Training-free' indicates no encoder training or diffusion model fine-tuning is needed. 
  `Single ref.' means the method can operate with only one reference image.
  }
  \label{tab:img_compare}
  \centering
 \small %
  \begin{tabular}{l p{1cm} p{1.5cm} p{0.9cm} }
    \toprule
    Name     &  Training free  & 
    Concept suppression  & Single ref. \\
    \midrule
    IP-Adapter~\citep{ye2023ipa} &  \xmark  
    & \cmark  & \cmark     \\
    Consistory~\citep{tewel2024training}    & \cmark      
    &  \xmark & \cmark \\
    Chosen one~\citep{avrahami2023chosen} & \xmark   
    &  \xmark & \xmark \\
    ELITE~\citep{wei2023elite}   & \xmark 
    & \xmark  &  \cmark \\
    BLIPD~\citep{li2024blip}    & \xmark
    &  \xmark    & \cmark  \\
    Ours & \cmark 
    & \cmark & \cmark \\
    \bottomrule
  \end{tabular}
  \normalsize %
\end{wraptable}
\section{Experiments}
\label{others}

We conduct experiments to show that \ours ~can help control consistency in two important tasks: image generation and video generation.

\vspace{-0.2cm}
\subsection{Controllable consistency in \\ image generation}

We use a fine-tuned SDXL of higher quality, ProtoVision-XL, as the base model for our experiments. For simplicity, we will refer to it as SDXL hereafter. We have replaced all the self-attention layers in SDXL with \rma, using the first sample in the batch as the reference image.

\vspace{-0.1cm}
\paragraph{Evaluation baselines}
In this section, we compare \ours ~with several baseline approaches: (1) SDXL~\cite{podell2023sdxl} without any modifications to its architecture; (2) encoder-based methods, such as IP-Adapter~\cite{ye2023ipa} and BLIPD~\cite{li2024blip}. For encoder-based methods, we initially generate a reference image using SDXL and then utilize this image as input. Additionally, we present a comparison of several other methods in \cref{tab:img_compare}.

\subsubsection{Consistent image generation}\label{sec:consist}
\begin{figure}[h]     
	\centering
    \vspace{-0.3cm} 
      \begin{subfigure}{0.49\textwidth}
    \centering
    \includegraphics[width=1\linewidth , align = c]{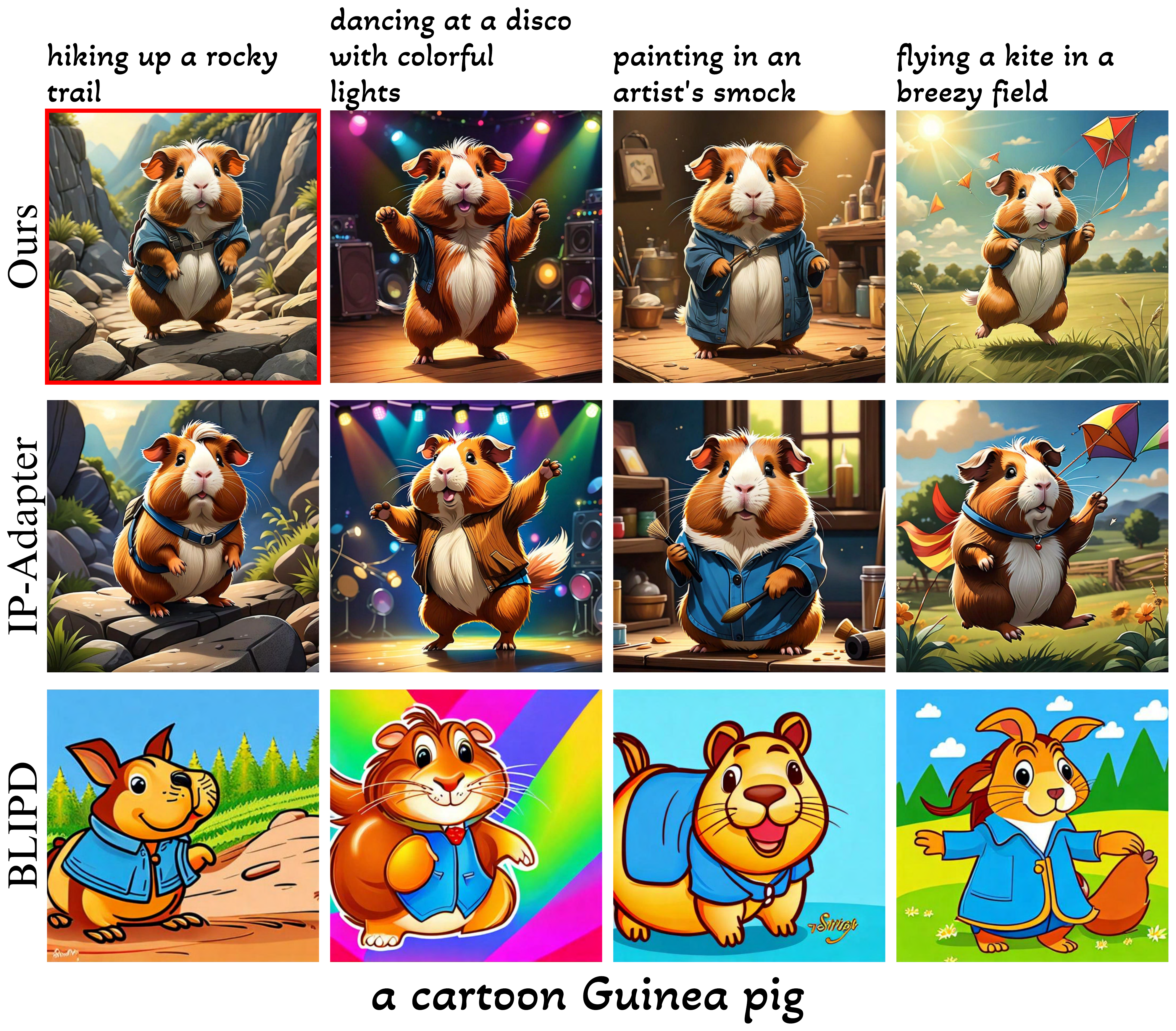}
  \end{subfigure} 
  \begin{subfigure}{0.485\textwidth}
    \centering
\includegraphics[width=1\linewidth, align = c]{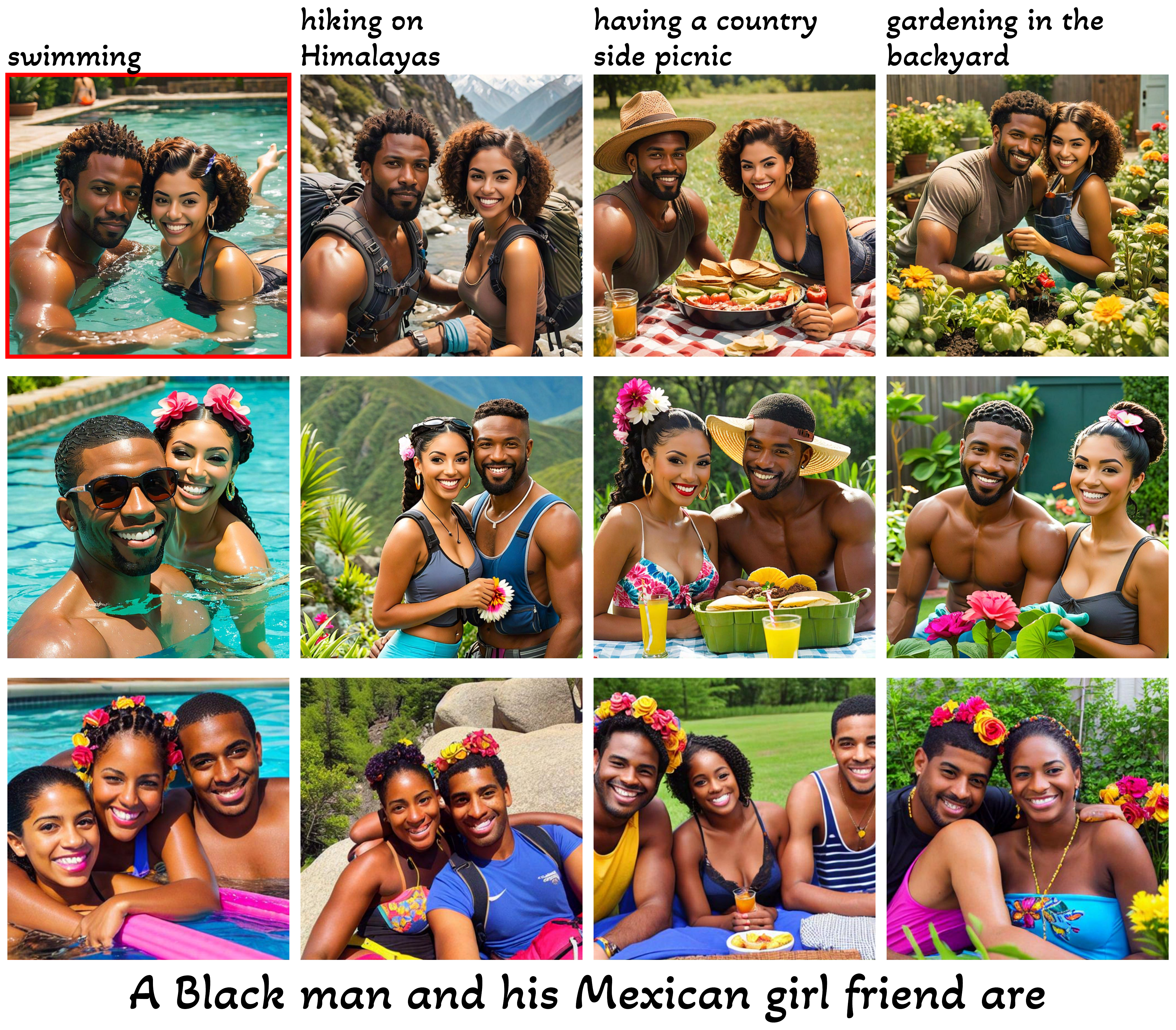}
  \end{subfigure}    
    \caption{
    The {\color{darkred}reference} image for all methods is framed in {\color{darkred}red}. Our method tends to produce more consistent outfits, hairstyles, and facial features compared to IP-Adapter and BLIPD. The visual quality of BLIPD is not comparable, as it utilizes SD1.5~\citep{rombach2022highresolution} as its base model.
    }  \label{fig:comp1}
\end{figure} 
For this task, we use 
$c \in [0.3,0.4]$ for our method.
We show qualitative results in \cref{fig:comp1}. 
IP-Adapter establishes a strong baseline,  especially on single subject consistent generation. However, it requires additional computational resources and data for training the image encoder compared to our approach.
BLIPD underperforms in terms of both visual quality and consistency relative to~\ours.
Additionally, we find that the concatenated attention used in studies like \citet{tewel2024training} and \citet{zhou2024storydiffusion} produces results that are quite similar to our \rma~mechanism, demonstrating that \rma~is capable of replicating the effects of concatenated attention. We also have the flexibility to incorporate other techniques from these studies~\citep{tewel2024training,zhou2024storydiffusion} to enhance pose diversity and reduce background leakage.
For \textbf{multi-subject} consistent generation, we find \ours ~can straightforwardly work for semantically different objects 
even without using separate subject masks.
This observation aligns with ConsiStory~\citep{tewel2024training}.
Further comparative results are available in the supplementary material.

\subsubsection{Blend features from multiple images}\label{sec:blend}

\begin{wrapfigure}[17]{r}{0.5\textwidth}
\vspace{-1.5cm}
\centering
    \includegraphics[width=1\linewidth , align = c]{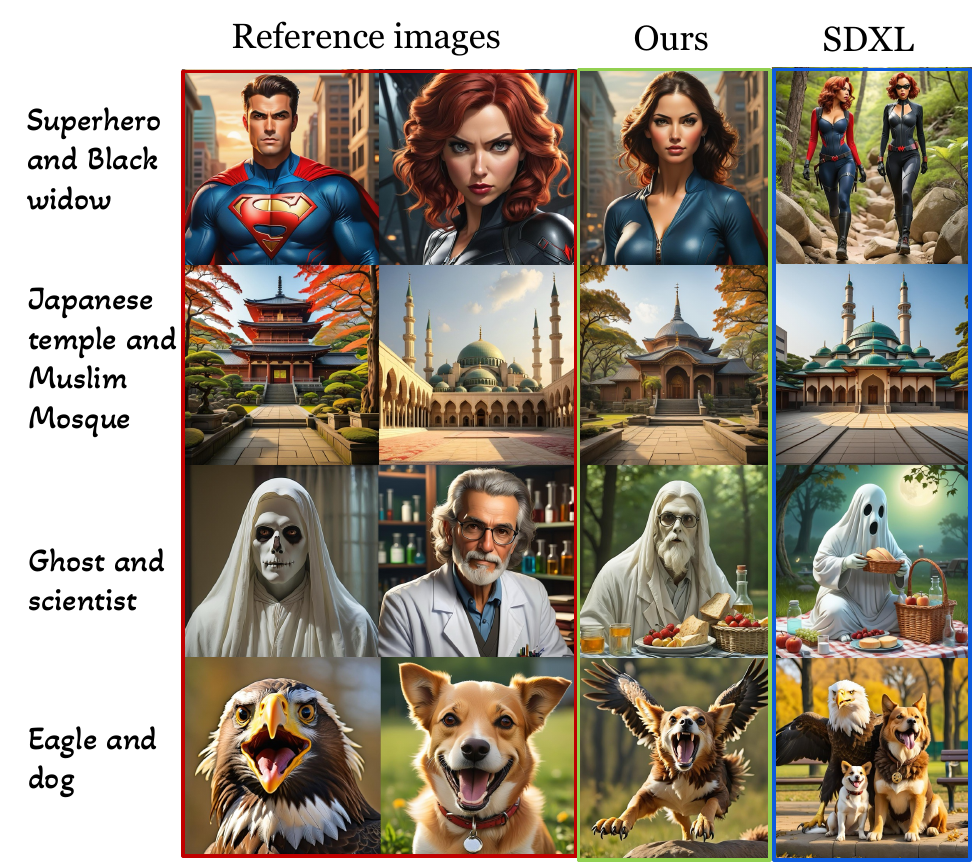}   
    \caption{\textbf{Multiple Reference Images:} The {\color{darkred}reference} images are highlighted with a {\color{darkred}red} frame, and the third image in each set is the resultant blended image. \ours~effectively assimilates features from the distinct reference images into a single and cohesive entity, demonstrating robust feature integration capability.  
    }  \label{fig:blend}
\end{wrapfigure} 
$\ours$ also supports the use of multiple reference images. In our implementation, we designate the first $N$ images in a batch as reference images. Features from these reference images are then incorporated into the subsequent images within the same batch through every self-attention layer. Formally, the extended \rma~with multiple references is defined as
\begin{align}
&X_{\rma}' = \sum_{j=1}^N c_j \cdot \text{Attention}(Q_i, K_j, V_j) + \nonumber \\
& (1 - \sum_{j=1}^N c_j) \cdot \text{Attention}(Q_i, K_i, V_i), \quad 
\label{eq:rma_multi}
\end{align}
for certain $ i > N$. Here, the attention mechanism ensures that the $i$-th image in the batch receives features from the first $1 \sim N$ reference images. 
In practice, we use $c_j \in [0.2,0.4]$ for any $j=1,\ldots, N$ in our method.
We demonstrate the capability of \ours ~to seamlessly blend distinct semantic features from two reference objects into a new object in \cref{fig:blend}. 
This task proves challenging when relying solely on prompt engineering. We attempted to achieve this task with SDXL using text prompts. For instance, if we aim to merge two objects, $\alpha$ and $\beta$, we might use prompts like ``an $\alpha$-like $\beta$'' or ``a $\beta$ in the style of $\alpha$.'' However, with such text prompts, SDXL either ignores the similarity with one of the reference images or frequently produces multiple objects instead of a single and cohesive entity. In \cref{fig:blend_3reference}, we show that \ours ~can blend \textit{three} distinct subjects: a dwarf, Black Widow, and Winnie the Pooh, encompassing a range of mythological being, human, and animal.

\subsubsection{Diverse image generation}\label{sec:diverse}

\begin{wrapfigure}[5]{r}{0.3\textwidth} %
\vspace{-1.7cm}
\centering
      \begin{subfigure}{0.14\textwidth}
    \centering
    \includegraphics[width=1\linewidth , align = c]{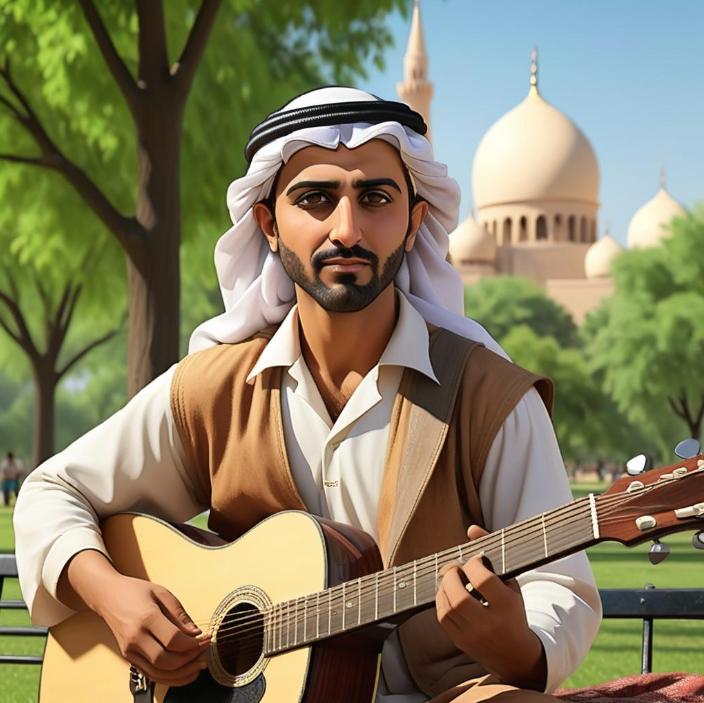}
  \end{subfigure} 
  \begin{subfigure}{0.14\textwidth}
    \centering
\includegraphics[width=1\linewidth, align = c]{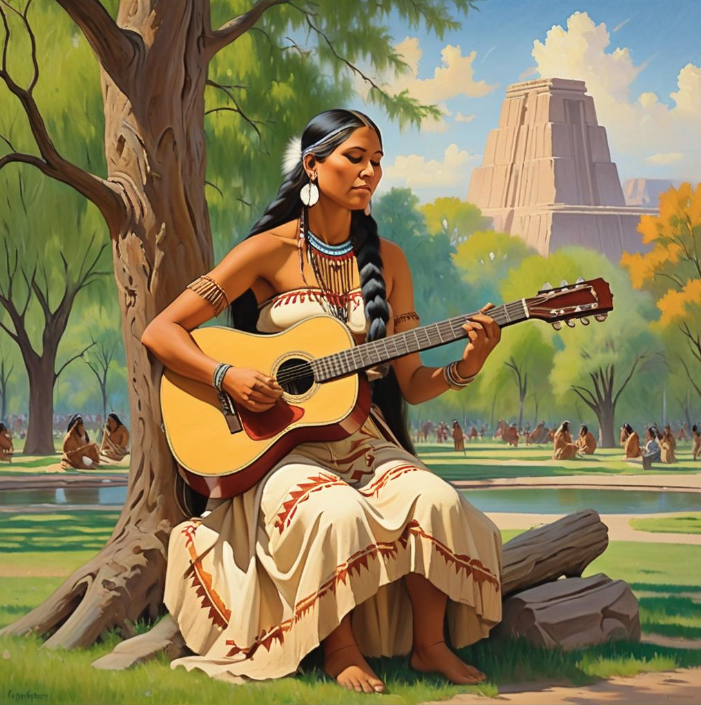}
  \end{subfigure}    
    \caption{Negative reference images for examples in \cref{fig:diverse_comp}.
    }  \label{fig:neg_anchor}
\end{wrapfigure} 

\begin{figure}[h]     
	\centering
      \begin{subfigure}{0.49\textwidth}
    \centering
    \includegraphics[width=1\linewidth , align = c]{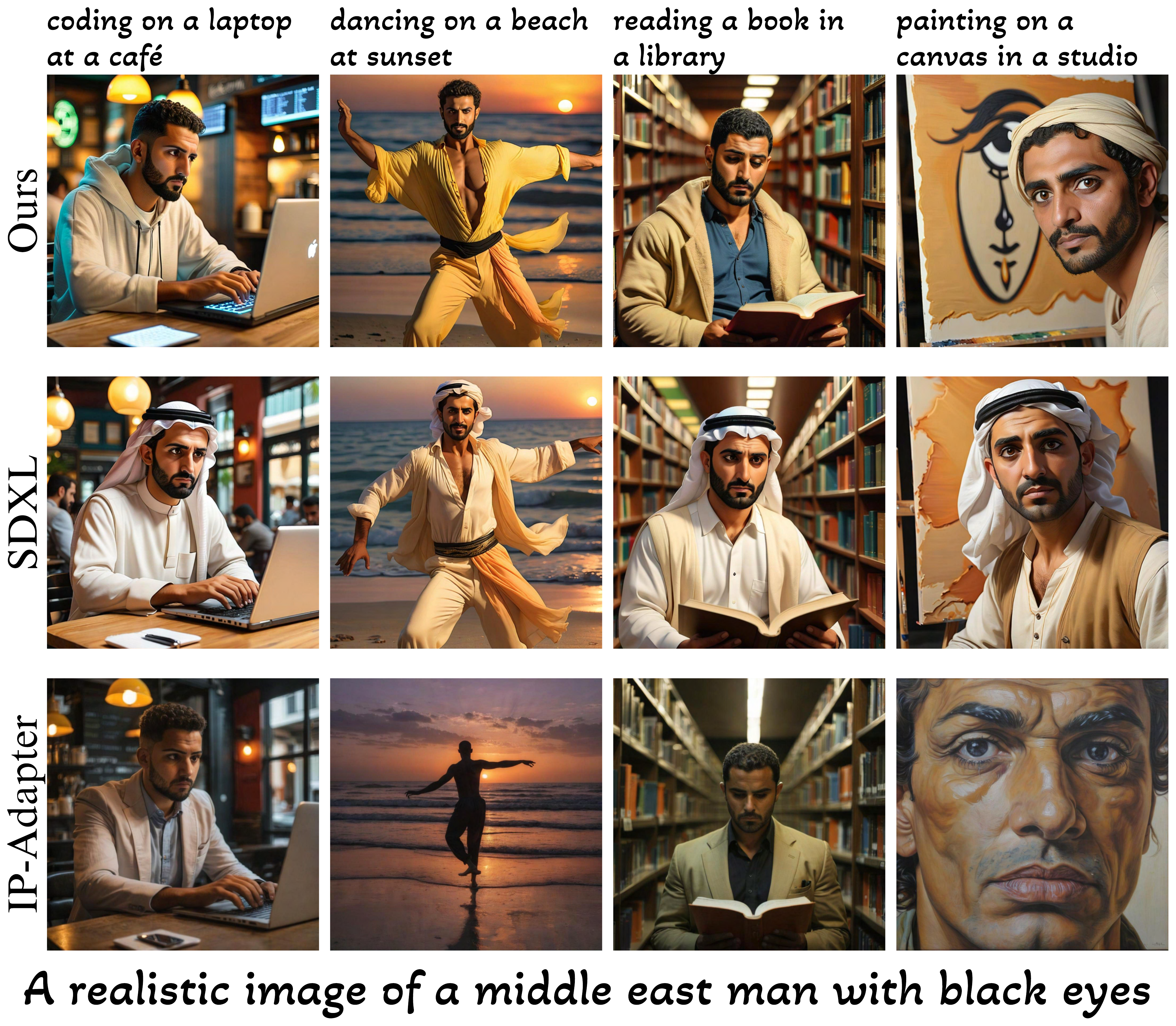}
  \end{subfigure} 
  \begin{subfigure}{0.485\textwidth}
    \centering
\includegraphics[width=1\linewidth, align = c]{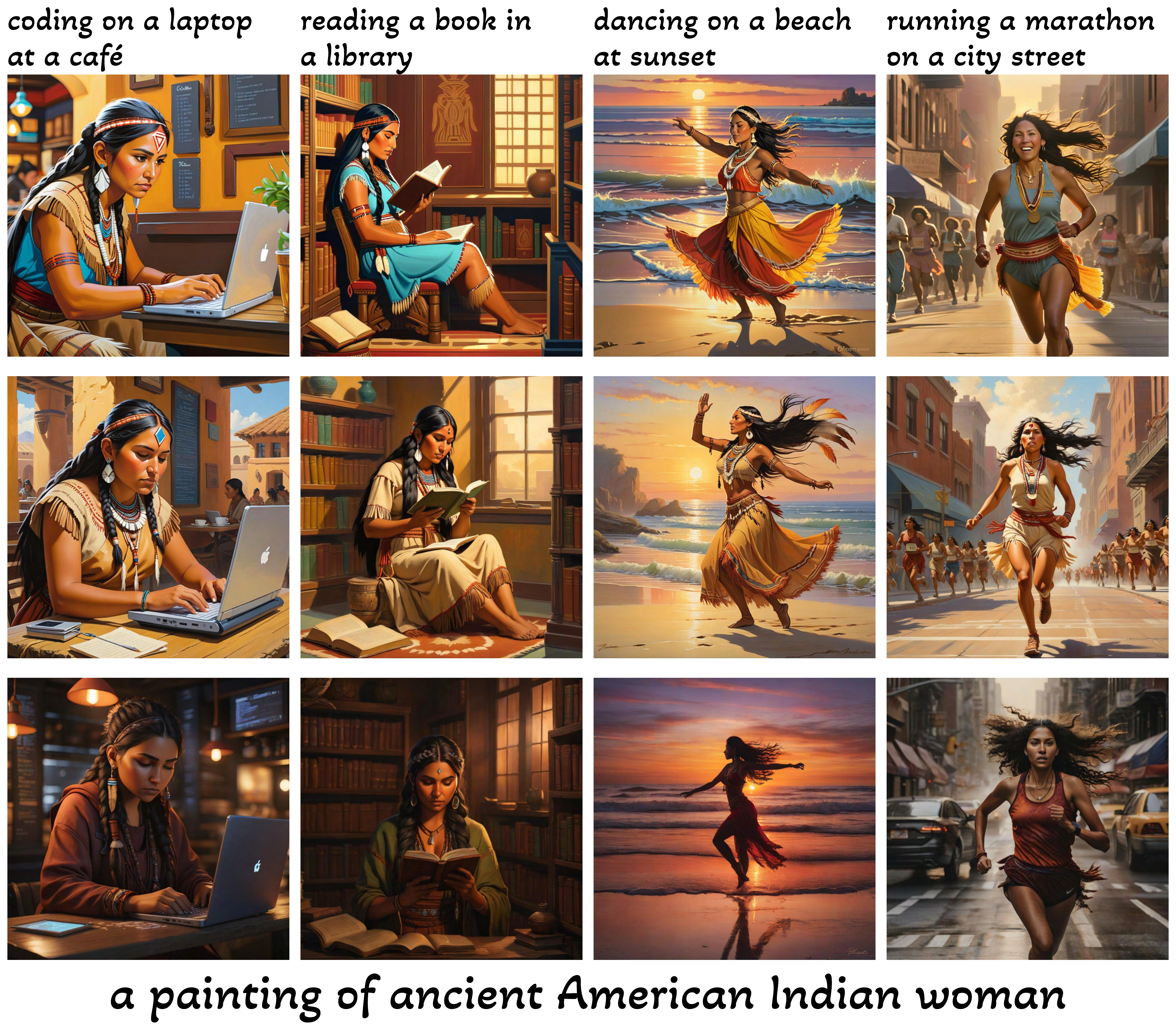}
  \end{subfigure}    
    \caption{{\color{Blue} Diverse image generation}: 
   Our method enhances diversity in outfits, hairstyles, and facial features, all while ensuring accurate text alignment. For example, while SDXL frequently generates headscarves in the first scenario and beige-colored clothes in the second, \ours ~can vary the presence of headscarves in the left example and produce clothing in different colors in the right example. Conversely, although IP-Adapter can create even more diverse images, it often fails to adhere to the style and human activity instructions in the text prompts. Additionally, it often produces overly small persons that lack detail.
    }  \label{fig:diverse_comp}
\end{figure}

Our method offers substantial flexibility in parameter tuning, enabling diverse image generation by setting the coefficient $c$ to a negative value. This feature is particularly valuable in addressing overfitting issues in image generation. 
For instance, when using SDXL to generate Middle Eastern faces, the output frequently includes similar headscarves, faces and outfits, as illustrated on the left side of \cref{fig:diverse_comp}.

In this task, we use $c = -0.3$ for our method.
We present a qualitative comparison in \cref{fig:diverse_comp}, with negative reference images displayed in \cref{fig:neg_anchor}. Upon comparing our method with IP-Adapter, we note that IP-Adapter may not adhere as closely to the text prompt. We attribute this to IP-Adapter's modification of cross-attention, which can impact text alignment. In contrast, our method focuses on modifying self-attention, thereby preserving the integrity of cross-attention and ensuring more accurate text alignment. We show additional quantitative results in \cref{fig:img_quant}.

\vspace{-0.3cm}
\subsection{Improving temporal-consistency in video generation}
Not only can we apply \rma ~to T2I generation, but it also effectively stabilizes video generation, where flickering commonly degrades quality. This section shows that using the first generated frame as a reference can greatly improve video generation. By injecting its features into the spatial self-attention layers of subsequent frames with a reference strength of 
$c=0.2$, we significantly stabilize these frames and enhance the temporal consistency of VDM.

We employ \texttt{SVD-img2vid-xt-1-1} as our I2V base model. Technically, our approach is 
compatible with any VDM, but we choose SVD as it is the best open-source model available. Although this model usually produces consistent videos from visually perfect images, we have noted that minor, often imperceptible flaws in the input images can significantly degrade the quality of the generated videos. Our method effectively stabilizes video quality in these scenarios.

\subsubsection{Temporal-consistent video generation}\label{sec:consist_video}

\begin{figure}[h]     
\centering
\vspace{-0.2cm}
  \begin{subfigure}{0.5\textwidth}
    \centering
    \includegraphics[width=1\linewidth , align = c]{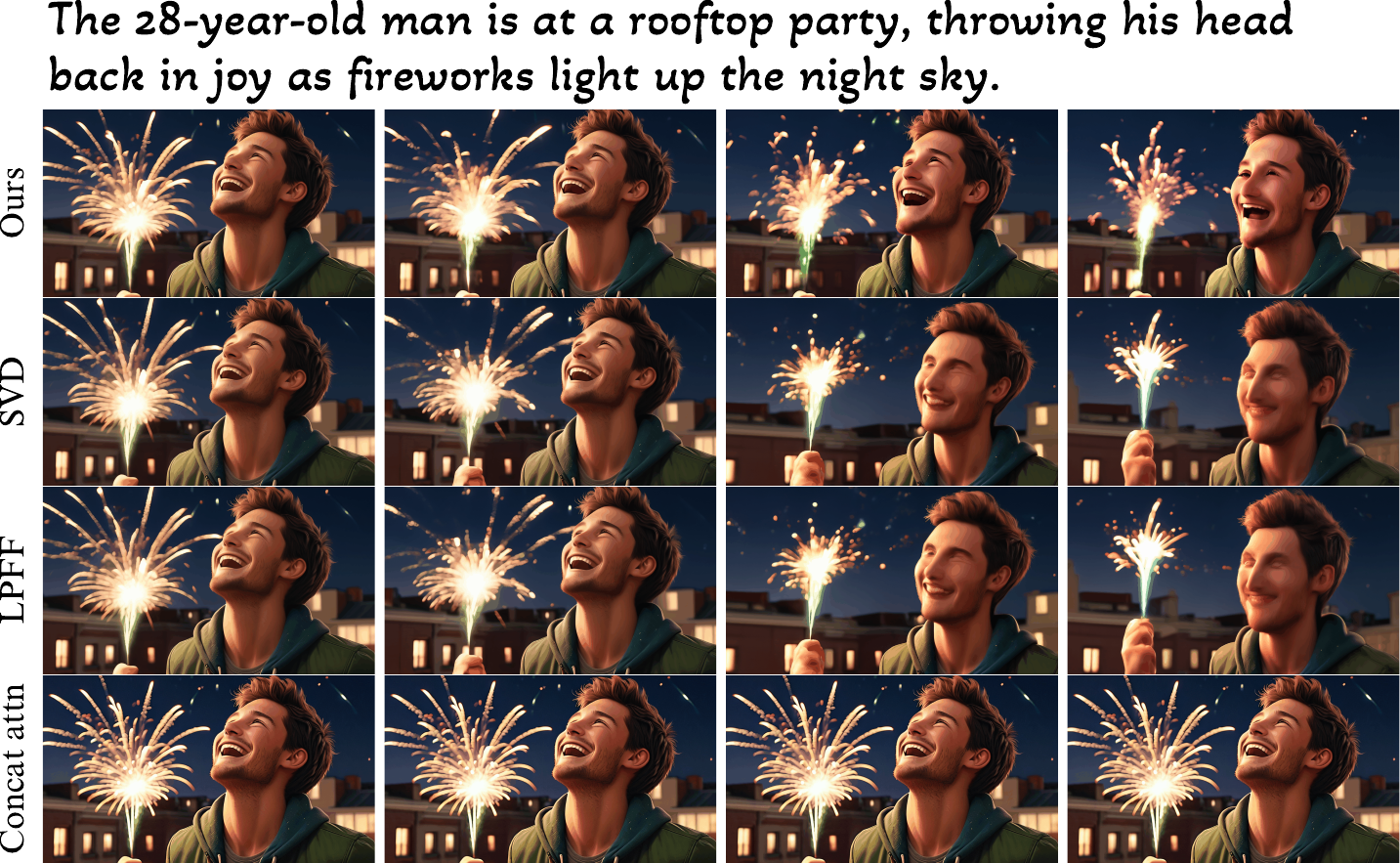}
  \end{subfigure} 
  \begin{subfigure}{0.49\textwidth}
    \centering
\includegraphics[width=1\linewidth, align = c]{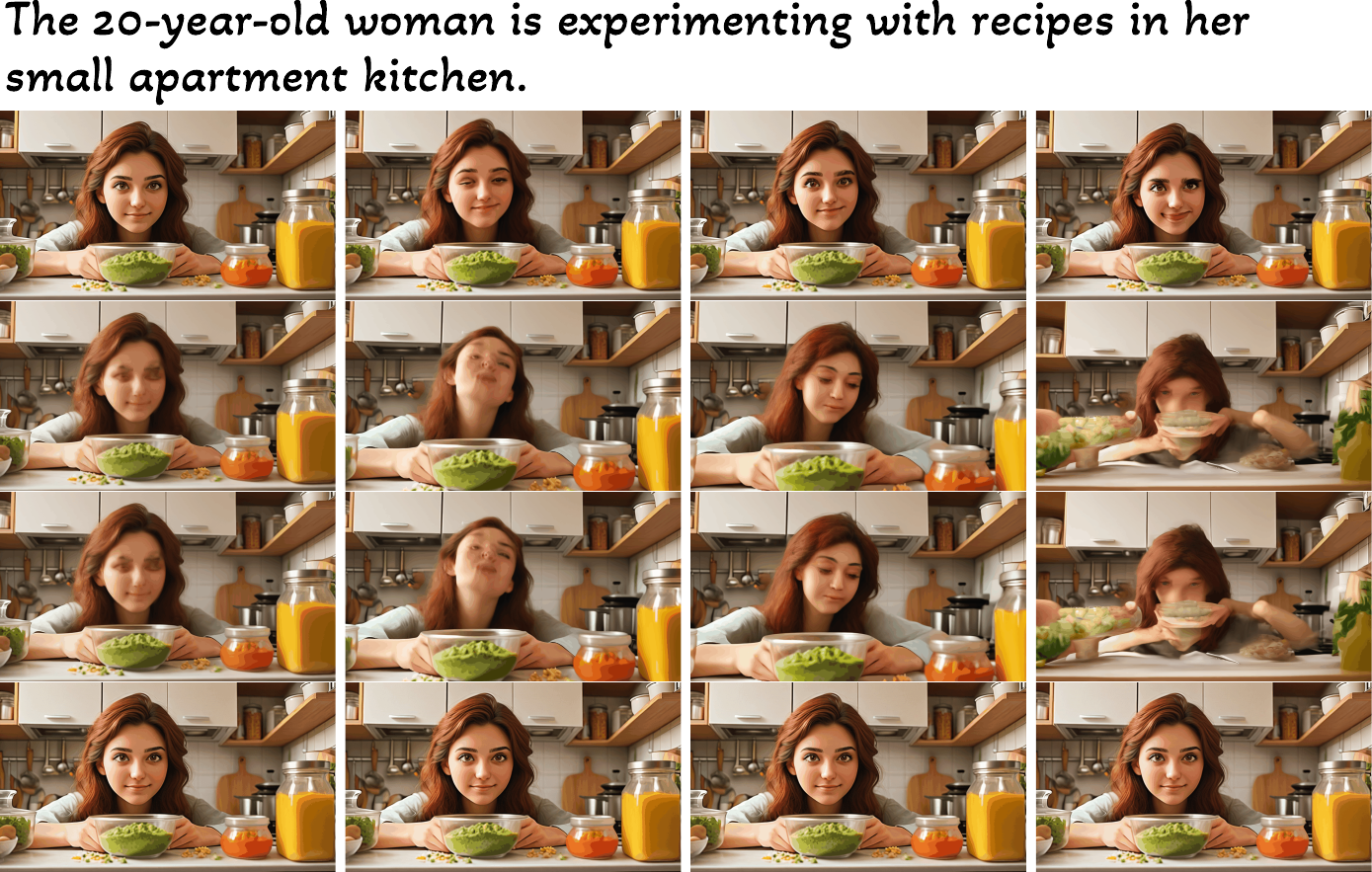}
  \end{subfigure}    
    \caption{Comparison of training free techniques to improve temporal consistency in video generation.
    }  \label{fig:consistent_vid}
\end{figure} 

In this part, we use the SDXL model to generate an image from a prompt and then pass this image to the SVD model. For \textbf{evaluation baselines}, we compare \ours ~with several training-free methods: unmodified SVD, Cross-Frame Attention~\citep{khachatryan2023text2video}, Concatenated Attention~\citep{wu2023tune}, and Temporal Low Pass Frequency Filter (LPFF)~\citep{zhang2024videoelevator}. Temporal LPFF is arguably superior to Spatial-Temporal LPFF by \citet{wu2023freeinit}, which shows Spatial-Temporal LPFF can result in blurry frames. We evaluate the Temporal LPFF using a fast sampling method that avoids the computationally intensive process of iteratively performing backward and forward diffusion at each denoising step.

The visualization results are displayed in \cref{fig:consistent_vid}. We observe that both Cross-Frame Attention and Concatenated Attention result in completely static videos, whereas LPFF shows minimal improvement. Our method proves to be the most effective in preventing flickering while preserving motion.

\subsubsection{Stabilizing personalized video generation}\label{sec:pv}
Finally, we explore the application of \ours ~to personalized video generation. Inspired by \citet{ku2024anyv2v}, starting with an image of a person, we use InstantID~\citep{wang2024instantid} to generate a personalized initial frame. This frame is then fed into SVD to create a short video. However, we observe that using the output from InstantID for SVD generation leads to a significantly higher failure rate compared to using the initial frame generated by SDXL. We attribute this increased failure rate to InstantID's propensity for producing images with more flaws, such as overly saturated colors, and distorted limbs, highlighting the potential demand for \ours ~in this task.

\begin{figure}[h!]
  \centering
    \makebox[\linewidth]{\includegraphics[width=1\linewidth]{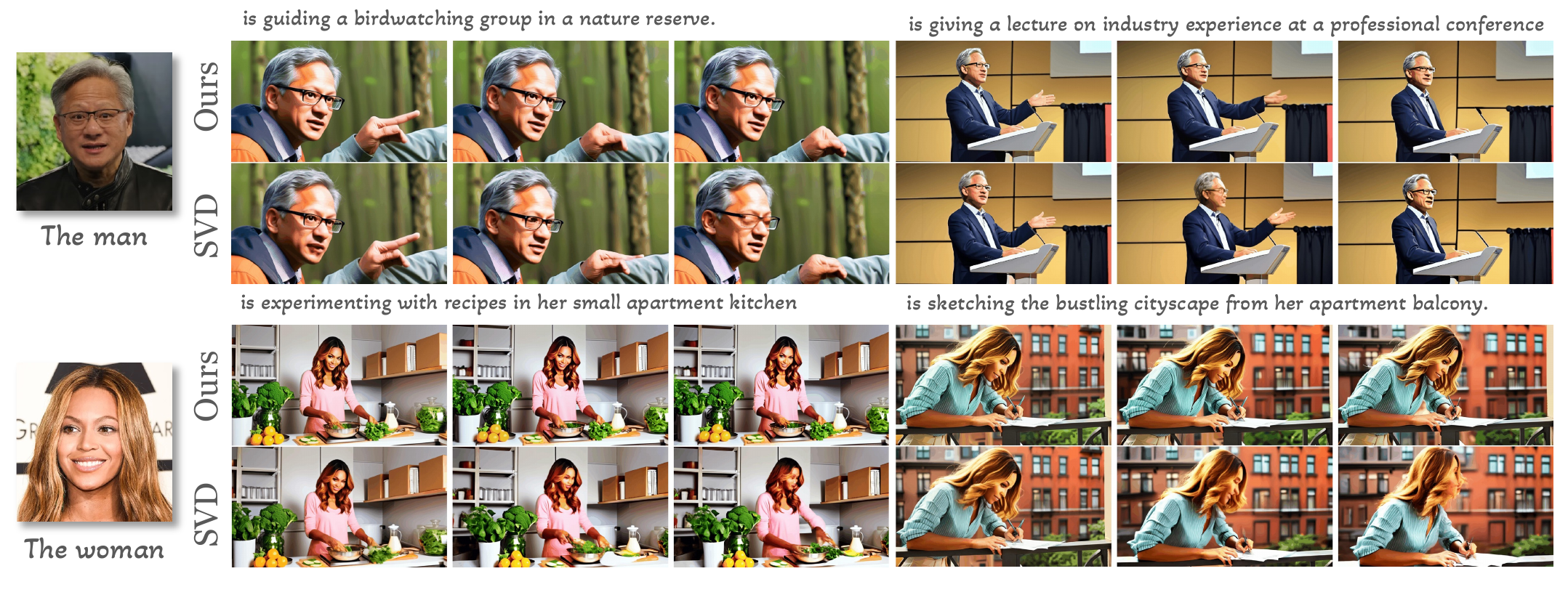}}
  \caption{By injecting the features of the first frame into the generation of subsequent frames, \ours ~reduces flickering and facial distortions. The additional videos can be viewed \href{https://refdrop-anonymouspaper-f37a6c745f264e0ff8b994669d71e9ca5f34d07a.gitlab.io/index.html}{here}.
  }
  \label{fig:person_vid}
\end{figure}

Several other methods are available for personalized video generation, as described in works \citep{wang2024customvideo,jiang2023videobooth,ma2024magic,Hu2023-wr}. 
Our method, which is designed to enhance temporal consistency, can be integrated with some of these existing approaches.
For example, in the case of Magic-Me \citep{ma2024magic}, 
our attention mechanism \rma ~ can be incorporated into their AnimateDiff \citep{guo2023animatediff} backbone. 
For the evaluation in this section, we primarily focus on comparisons with naive SVD generation, as it directly relates to our goal of enhancing temporal consistency.

We present such comparison 
between our \ours ~enhanced generation to the naive SVD generation
in \cref{fig:person_vid}. \ours ~effectively preserves identity during video generation, offering improvements similar to those achieved by increasing the CFG. However, unlike increasing CFG, which often results in over-saturation of videos, our approach does not produce such artifacts. 
We present additional automatic metrics in \cref{tab:comparison_metrics} to show that \ours~can enhance the quality of the generated videos.

\section{Human evaluation}
\begin{figure}[h!]
  \centering
    \centering
    \vspace{-0.3cm}
    \includegraphics[width=1\linewidth]{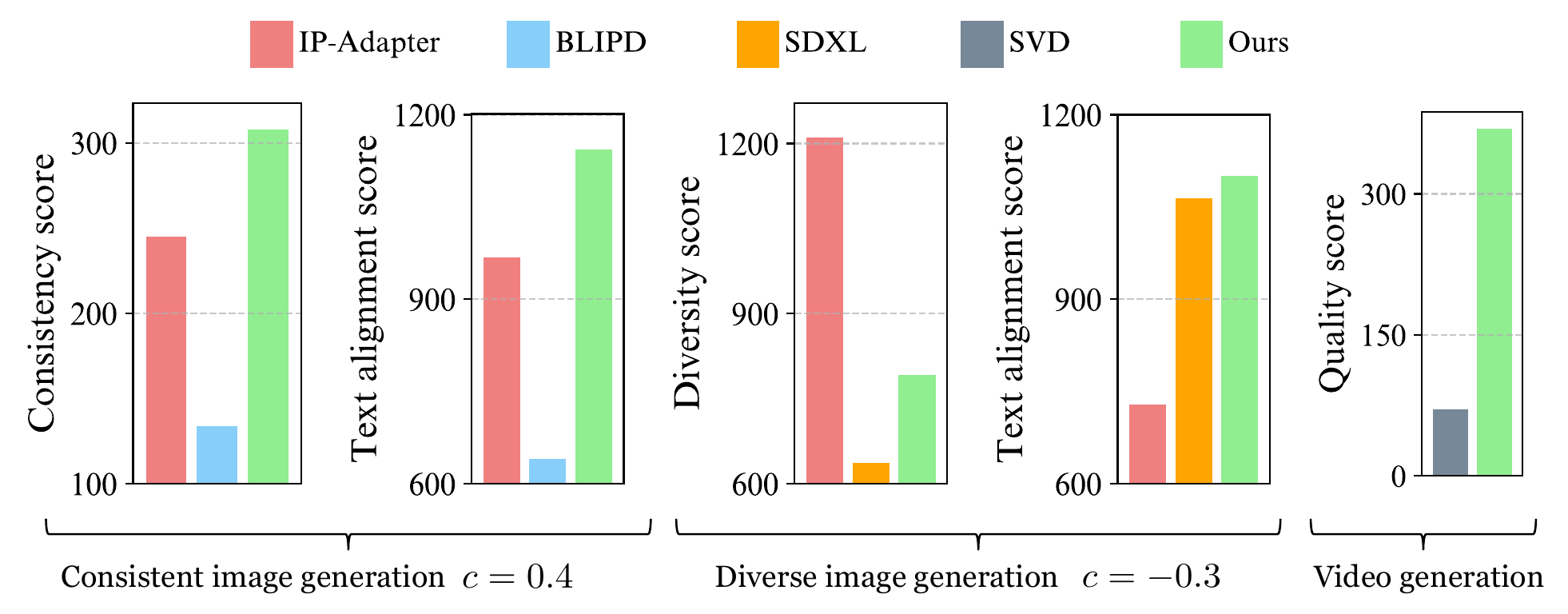}
  \caption{A higher score indicates a superior result. In the category of consistent image generation, participants showed a preference for \ours, with IP-Adapter ranking slightly behind. In diverse image generation, while IP-Adapter was favored for its variety, it significantly compromised text alignment. Conversely, \ours ~maintained a good balance, achieving diversity while preserving text alignment. In personalized video generation, users clearly preferred our approach, demonstrating substantial improvements over the SVD results.} \label{fig:human}
  \end{figure}
We conducted a human evaluation study using Google Forms. Our survey is structured into three distinct categories: 1) Consistent Image Generation, 2) Diverse Image Generation, and 3) Personalized Video Generation. Initially, we utilized ChatGPT to generate text prompts, then processed approximately 100 small tasks per category using both baseline methods and our approach. From these, we randomly selected 10 sets for evaluation. In the first two categories, participants assessed the consistency and diversity of the images, as well as text alignment. For the third category, participants were asked to select the video with better quality. The aggregated scores are detailed in \cref{fig:human}. We collected responses from 44 distinct users in total.
More details appear in \cref{sec:user_study}.

\section{Conclusion}\label{sec:conclude}
In this study, we propose a method that effectively uses one or multiple generated images to guide the generation of other images or video frames. Through extensive experiments, our method has proven useful for flexible consistency control in image generation and has improved temporal consistency in video generation. In particular, we show applications in consistent and diverse image generation, feature blending from multiple images, and enhancement of video temporal consistency.
Moreover, our approach is versatile on network architecture as it applies not only to UNet-based models but also to transformer-based diffusion models like DiT~\citep{peebles2023scalable}.  

Looking ahead, several promising avenues for further research emerge from this study. Firstly, our experiments have not yet explored the use of attention masks; investigating their potential for precise control in image generation presents a compelling opportunity for future work. Another exciting prospect involves enhancing our method to accept clean reference images as input, similar to the IP-Adapter and other image personalization techniques. Achieving this capability would represent a significant advancement, particularly if coupled with  an optimal image inversion method.

\section{Limitation} \label{sec:limit}
In consistent image generation, our model sometimes struggles to exactly replicate the appearance of specific objects, tending to hallucinate rather than accurately reproduce items like lotion bottles.

\bibliography{ref}
\bibliographystyle{tmlr}

%% file: supple.tex
\newpage
\appendix

\section{Broader impacts}\label{sec:impacts}

The broader impacts of advancements in consistent character generation and personalized video generation extend across multiple domains, notably enhancing both creative and technological landscapes. In the media and entertainment industries, for instance, these methods can revolutionize character design, fostering more reliable representations. However, there is also a potential risk associated with our method, as it could be used to create fake profiles, highlighting the need for careful consideration of its applications.

\section{Relationship to Concatenated attention}\label{sec:relation_concat}
\begin{wrapfigure}[12]{r}{0.3\textwidth} %
\vspace{-1.3cm}
\centering
      \begin{subfigure}{0.12\textwidth}
    \centering
    \includegraphics[width=1\linewidth , align = c]{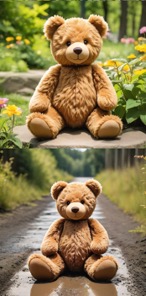}
    \caption*{Eq. \eqref{eq:matrix_coeff}-\eqref{eq:vector_coeff}}
  \end{subfigure} 
  \begin{subfigure}{0.15\textwidth}
    \centering
\includegraphics[width=1\linewidth, align = c]{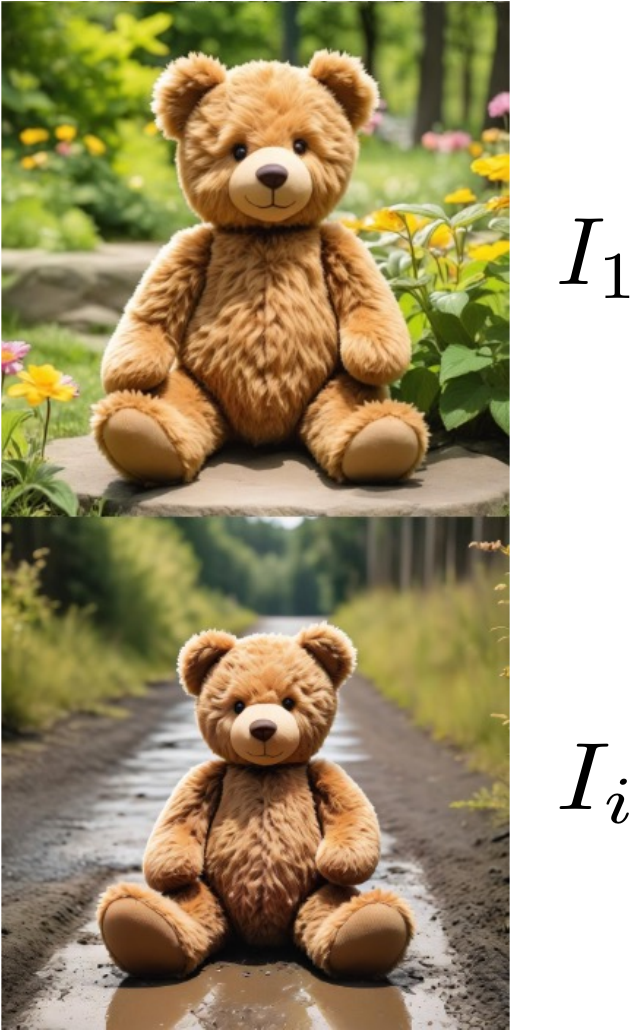}
\caption*{Concat attn~\eqref{eq:concat}}
  \end{subfigure}    
    \caption{Concatenated attention is our special case.
    }  \label{fig:concat_euiv}
\end{wrapfigure} 
In this section, we delve into details on how our \rma~framework can replicate the concatenated attention in \eqref{eq:concat}. We begin by noting the dimensions of the attention maps:

\begin{align}
    \text{Softmax}\left(\frac{Q_i K_1^\top}{\sqrt{d}}\right) & \in \mathbb{R}^{L \times L}, \\
    \text{Softmax}\left(\frac{Q_i K_i^\top}{\sqrt{d}}\right) & \in \mathbb{R}^{L \times L}, \\
    \text{Softmax}\left(\frac{Q_i [K_1; K_i]^\top}{\sqrt{d}}\right) & \in \mathbb{R}^{L \times 2L},
\end{align}
where \(L\) is the sequence length of the input hidden feature \(X\). 
We further define $\mathbf{1}_d$ as an all-ones vector of dimension $d$, and $\mathbf{1}$ as an all-ones matrix, sized appropriately to ensure the validity of the operations it is involved in.
To recover concatenated attention \eqref{eq:concat}, we extend the scalar \(c\) from \eqref{eq:rma} to a rank-1 weight matrix:
\begin{align}\label{eq:matrix_coeff}
C = \mathbf{c} \otimes \mathbf{1}_{d_v},
\end{align}
where all columns in \(C  \in \mR^{L \times d_v} \) are identical, represented by the vector \(\mathbf{c} \in \mR^{d_v} \), and \(d_v\) is the feature dimension of the value \(V\). We then transform the scalar dot product into a matrix element-wise product $\odot$, allowing \rma~to be expressed with this matrix coefficient as:
\begin{align}\label{eq:matrix_rma}
X_{\rma}' = 
C \odot \left( \text{Softmax}\left(\frac{Q_i K_1^\top}{\sqrt{d}}\right) V_1 \right)  +  (\mathbf{1} - C) \odot \left( \text{Softmax}\left(\frac{Q_i K_i^\top}{\sqrt{d}}\right) V_i \right).
\end{align}

Denote \(./\) and $\exp$ as the element-wise division and exponential operation respectively. By setting
\begin{align}\label{eq:vector_coeff}
    \mathbf{c} = \left({\exp}\left(\frac{Q_i K_1^\top}{\sqrt{d}}\right) \mathbf{1}_L \right) ./ \left({\exp}\left(\frac{Q_i [K_1; K_i]^\top}{\sqrt{d}}\right) \mathbf{1}_{2L}\right),
\end{align}
 \(X_{\rma}'\) can recover the concatenated attention~\citep{wu2023tune}
\begin{equation}
X_{\texttt{CAT}}' = \text{Softmax}\left(\frac{Q_i [K_1; K_i]^\top}{\sqrt{d}}\right) [V_1; V_i] = \text{Softmax}\left(\frac{ [Q_iK_1^\top; Q_iK_i^\top]}{\sqrt{d}}\right) [V_1; V_i]. 
\end{equation}
The reason for this to hold is simply the normalizing effect of softmax. The softmax operation would normalize each row in the attention map $\frac{Q_i [K_1; K_i]^\top}{\sqrt{d}}$ independently, thus our weight matrix is a rank-1 matrix with different rows.

Finally, we can also recover the Cross-Frame attention~\citep{khachatryan2023text2video} by setting the coefficient $c=1$.

\section{Additional results}

\subsection{Quantitative results}\label{sec:quant}

\begin{figure}[h]     
	\centering
      \begin{subfigure}{0.49\textwidth}
    \centering
    \includegraphics[width=1\linewidth , align = c]{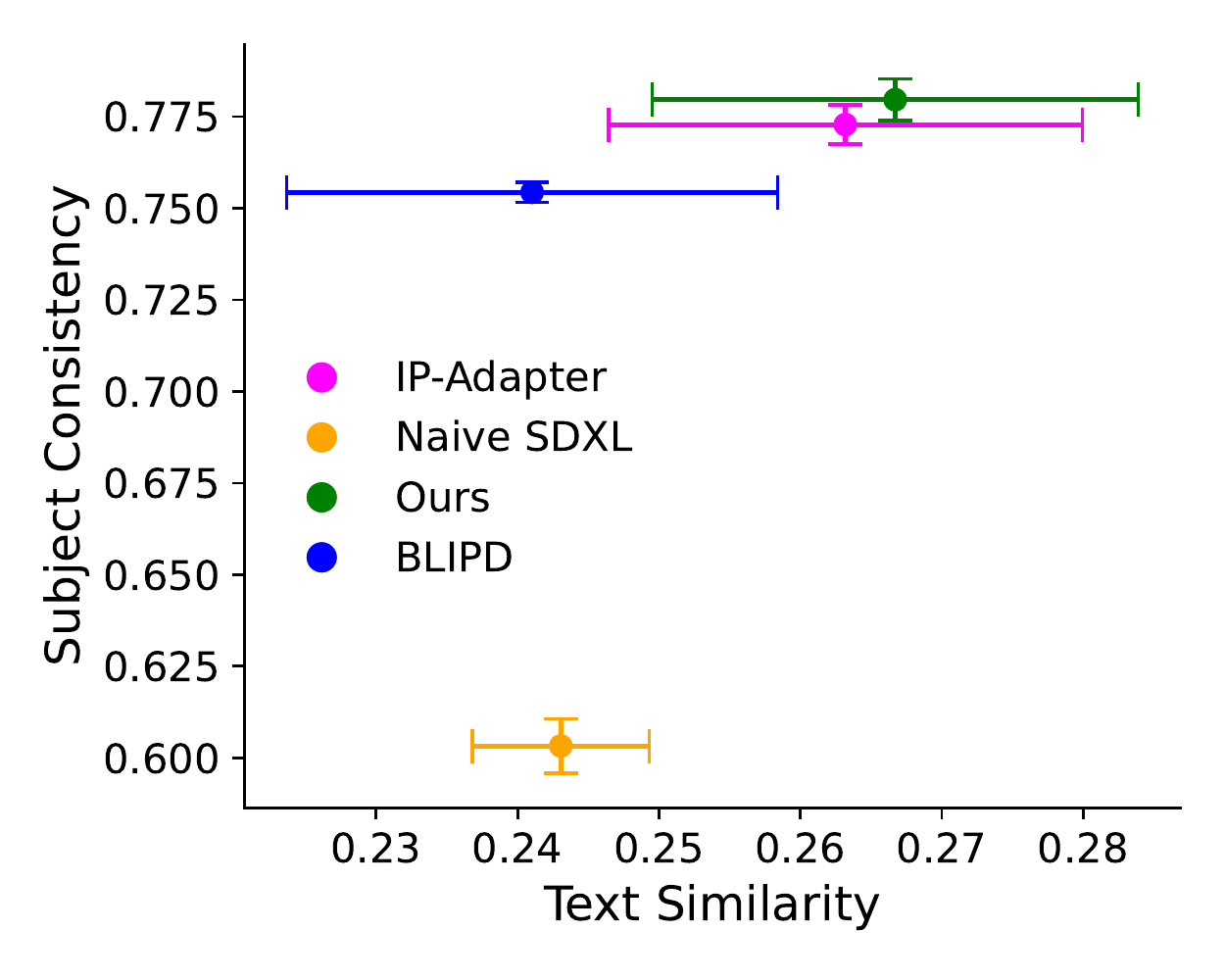}
  \end{subfigure} 
  \begin{subfigure}{0.485\textwidth}
    \centering
\includegraphics[width=1\linewidth, align = c]{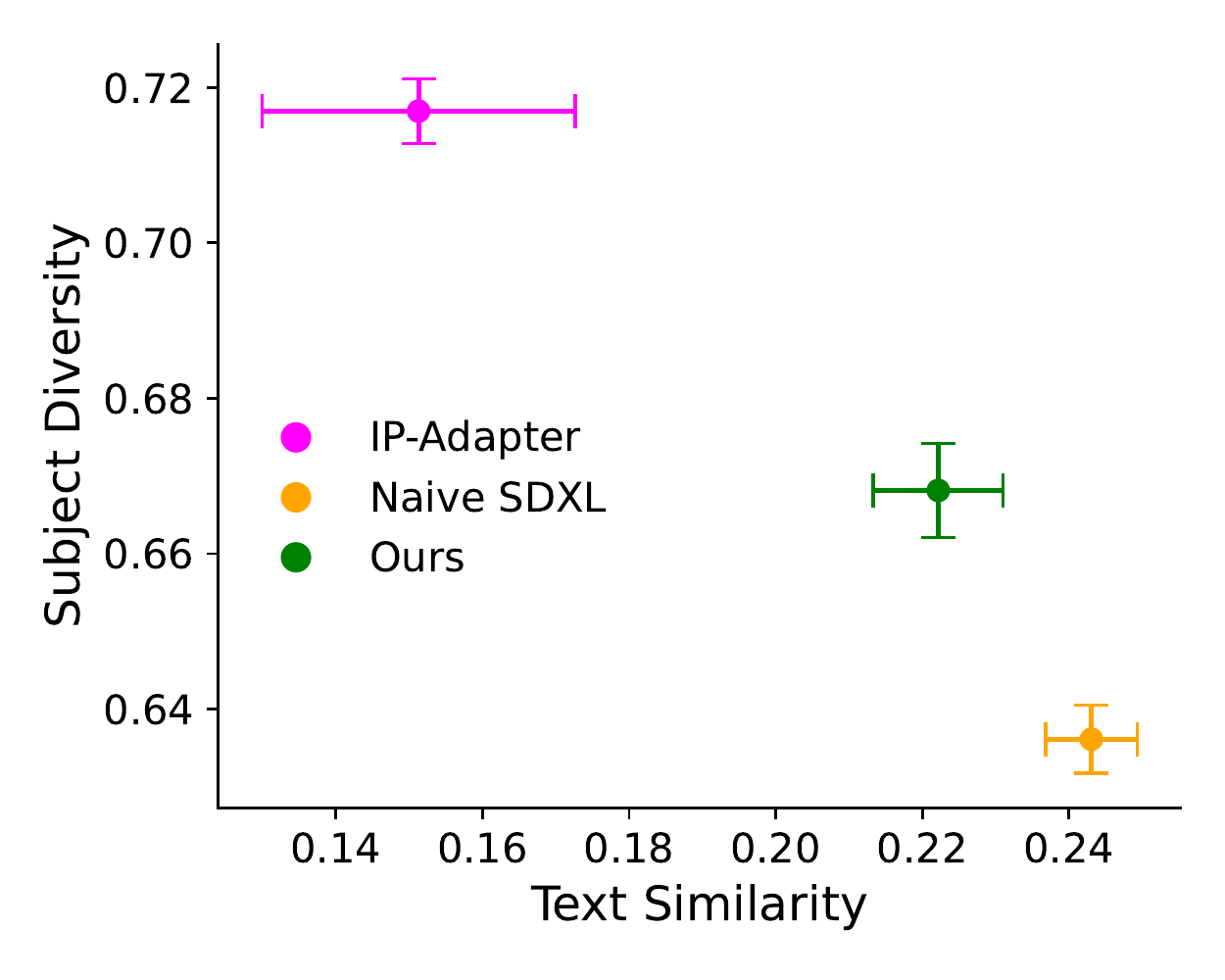}
  \end{subfigure}    
    \caption{
    Left: \textbf{Consistent Image Generation.} Our method achieves not only the highest subject consistency but also superior text alignment.
Right: \textbf{Diverse Image Generation.} Our approach maintains higher subject diversity with only a slight compromise in text alignment. In contrast, IP-Adapter exhibits the highest subject diversity but suffers from a significant reduction in text alignment. Error bars represent standard deviation.
    }  \label{fig:img_quant}
\end{figure}

\paragraph{Text-to-Image generation} We present quantitative metrics in Figure \ref{fig:img_quant}. Using the OpenCLIP model, \texttt{CLIP-ViT-g-14-laion2B}, we measure text-image similarity by averaging CLIP scores~\citep{hessel2021clipscore} across 100 pairs of text prompts and generated images. This measurement is repeated five times using different pairs for each method, and the variability is depicted through error bars. For assessing subject consistency, we utilize DreamSim~\citep{fu2023dreamsim}, after processing images to remove backgrounds\footnote{We use the Tracer-B7 model in  \\
\url{https://github.com/OPHoperHPO/image-background-remove-tool/?tab=readme-ov-file}} in order to focus analysis on foreground content. In tasks of diverse image generation, we employ LPIPS to gauge image diversity. We calculate the pairwise DreamSim or LPIPS distance between 400 image pairs per method, repeating these measurements with distinct pairs to ensure robust results, and report these findings with error bars. The measures of consistency and diversity are expressed as one minus the calculated DreamSim or LPIPS distances. These results demonstrate that our method is effectively situated on the Pareto-front, aligning with the human evaluations reported in Figure \ref{fig:human}.

\paragraph{Image-to-Video generation} 
We present a comparison of automatic metrics for video generation in \cref{tab:comparison_metrics}. All metrics are designed by EvalCrafter~\citep{liu2023evalcrafter}. Following EvalCrafter, we measure the quality of generated videos from four perspectives: overall quality, text alignment, temporal consistency, and motion quality.
Specifically, VQA$_A$ measures the aesthetic score, and VQA$_T$ evaluates common distortions such as noise and artifacts. CLIP Score quantifies the similarity between input text prompts and generated videos. 
For temporal consistency, we use CLIP-Temp to measure semantic consistency between frames, and also calculate face consistency, and warping errors. Finally, the flow score calculates the average optical flow across all video frames.
We generated 220 personalized videos using 220 distinct prompts for both SVD and \ours, utilizing images of four individuals shown in \cref{fig:person_vid}. 
The prompts included both close-up and distant descriptions. The metrics shown in \cref{tab:comparison_metrics} are averaged over these 220 videos. 
The statistics demonstrate that \ours ~reduces unnecessary flickering and improves overall quality. Surprisingly, we find that \ours ~not only improves the visual quality but also the text alignment.

\renewcommand{\arraystretch}{1.3}
\begin{table}[h!]
    \caption{
 Comparison of automatic metrics between SVD and \ours ~on video generation. An $\uparrow$ symbol indicates that higher values are better, while a $\downarrow$ symbol indicates that lower values are preferable.
 Our model shows improvements over the SVD base model in overall quality, text alignment, and temporal consistency. The flow score is the only metric where the SVD model scores higher, indicating more motion. However, the SVD model also exhibits greater jittering and flickering, as reflected in its larger warping error. Notably, a static video would register a flow score of zero. This suggests that our generated videos maintain a reasonable level of motion.
    }
    \label{tab:comparison_metrics}
    \vspace{0.3cm}
    \centering
    \begin{tabular}{c|cc|c|ccc|c}       
        \toprule
         & \multicolumn{2}{c|}{\textbf{Overall quality}} & \textbf{Text alignment} & \multicolumn{3}{c|}{\textbf{Temporal Consistency}} & \textbf{Motion} \\ 
        \hline         
         & VQA$_A$ $\uparrow$ & VQA$_T$ $\uparrow$ & CLIP score $\uparrow$ 
         & \begin{tabular}[c]{@{}c@{}}CLIP\\ Temp\end{tabular} $\uparrow $  & \begin{tabular}[c]{@{}c@{}}Face\\ consis.\end{tabular} $\uparrow $  & \begin{tabular}[c]{@{}c@{}}Warping\\ error\end{tabular} $\downarrow$  & \begin{tabular}[c]{@{}c@{}}Flow\\ score\end{tabular} $\uparrow$ \\ 
        \hline
        Ours & \textbf{94.27} & \textbf{89.91} & \textbf{20.84} 
        & \textbf{99.91} & \textbf{99.46} & \textbf{0.0058} & 2.62  \\        
        SVD & 93.25 & 86.20 
        & 20.76 & 99.83 & 99.20 & 0.0077 & \textbf{5.80} \\
        \bottomrule
    \end{tabular}
\end{table}

\subsection{Qualitative results}

\begin{figure}[h!]
  \centering
    \makebox[\linewidth]{\includegraphics[width=1\linewidth]{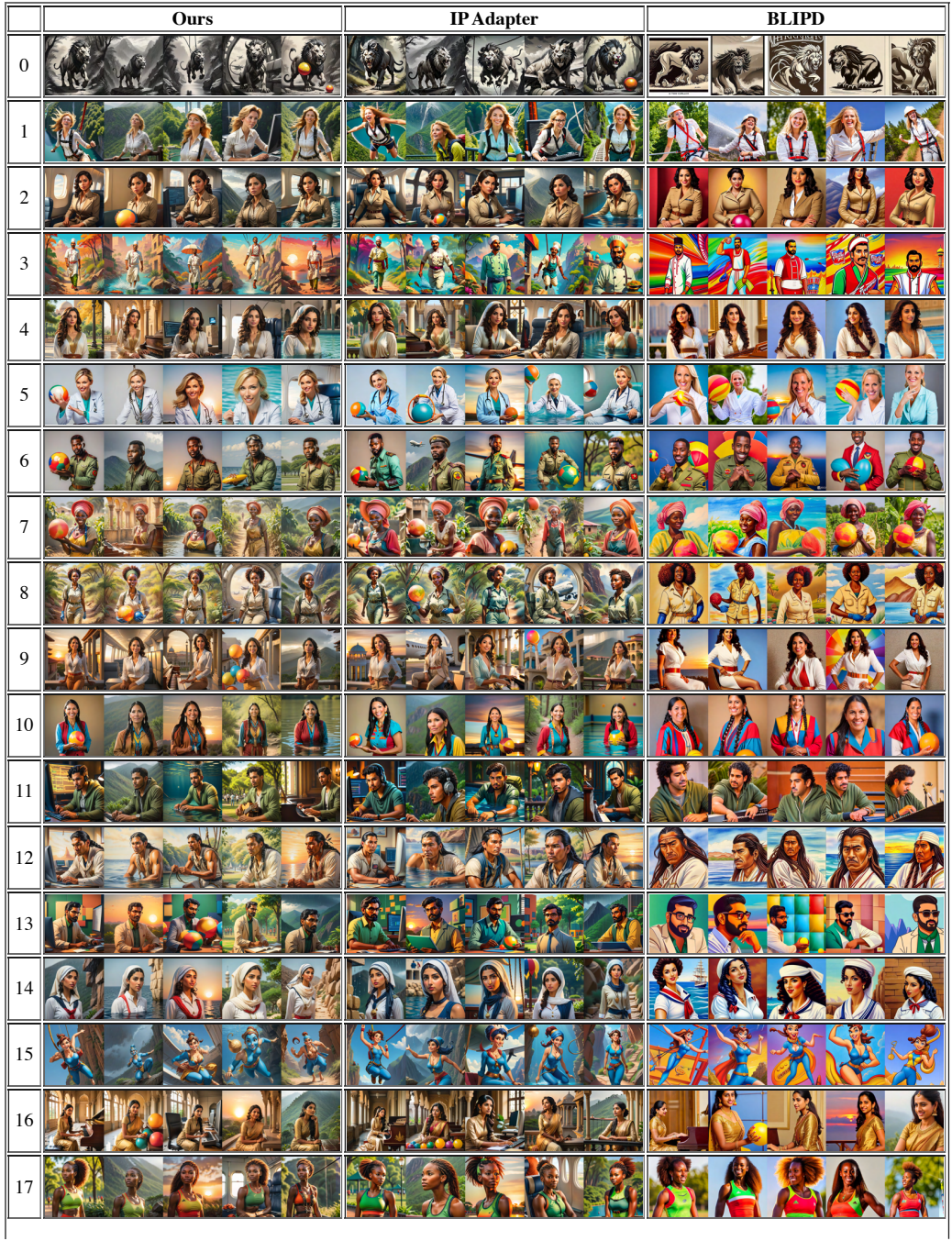}}
  \caption{More consistent image generation comparison.
  We observe that our \rma ~tends to produce generated individuals with similar poses in certain instances. To mitigate this issue, we can adopt strategies such as "using vanilla query features" and "self-attention dropout," as suggested by \citet{tewel2024training}.
  }
  \label{fig:consist_supple}
\end{figure}

\begin{figure}[h!]
  \centering
    \makebox[\linewidth]{\includegraphics[width=1\linewidth]{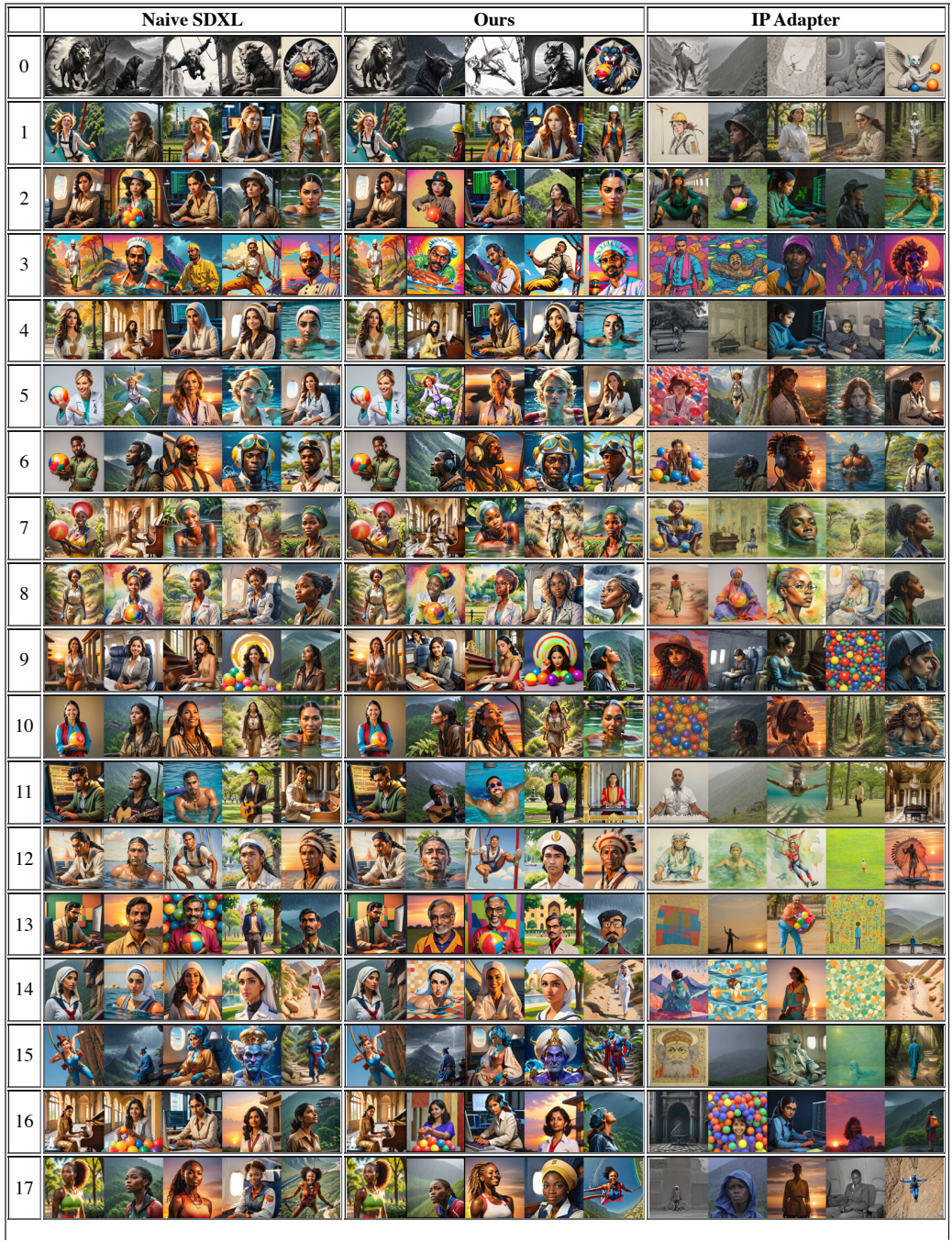}}
  \caption{More diverse image generation comparison.}
  \label{fig:diverse_supple}
\end{figure}

\textbf{Consistent and Diverse Image Generation:} We give more visualizations for consistent and diverse image generation in \cref{fig:consist_supple} and \cref{fig:diverse_supple}.
We attached images their original quality in \url{consistent_generation.pdf} and \url{diverse_generation.pdf} in the supplementary material.

\begin{figure}[h!]
  \centering
    \makebox[\linewidth]{\includegraphics[width=1\linewidth]{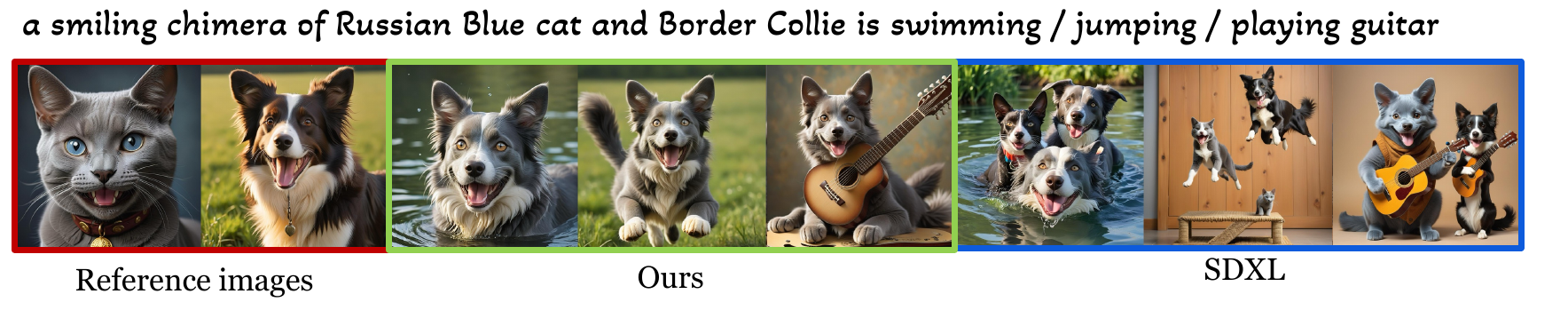}}
  \caption{Blending a dog and a cat in various activities: \ours ~successfully combines features from two reference images and closely follows the text prompt, whereas SDXL struggles to generate a single cohesive object even with the guidance from the text prompt.
  }
  \label{fig:blend2}
\end{figure}

\begin{figure}[h!]
  \centering
    \makebox[\linewidth]{\includegraphics[width=1\linewidth]{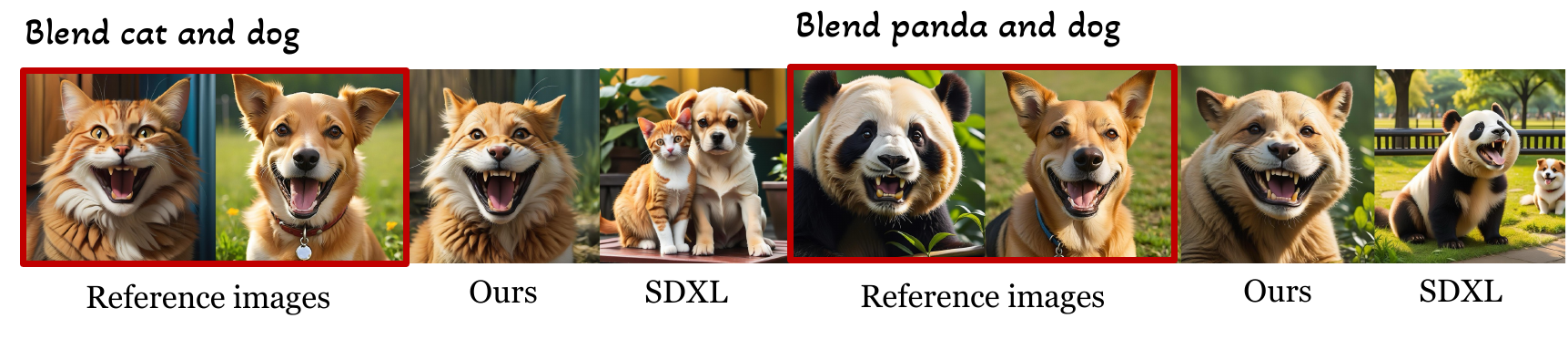}}
  \caption{More visualizations for blending two distinct animals. 
  One crucial strategy for our method to effectively blend two objects is to avoid explicitly naming them in the text prompt. We have discovered that using a generic term like "an animal" leads to better results than specifying "a cat-like dog." This trick minimizes the overly strong influence that explicit names can have, facilitating a more effective merger of the two subjects. For SDXL, we use the prompt "a chimera of [animal A] and [animal B]", but it  fails to generate a single and cohesive entity.
  }
  \label{fig:blend3}
\end{figure}

\begin{figure}[h!]
\centering
    \includegraphics[width=1\linewidth , align = c]{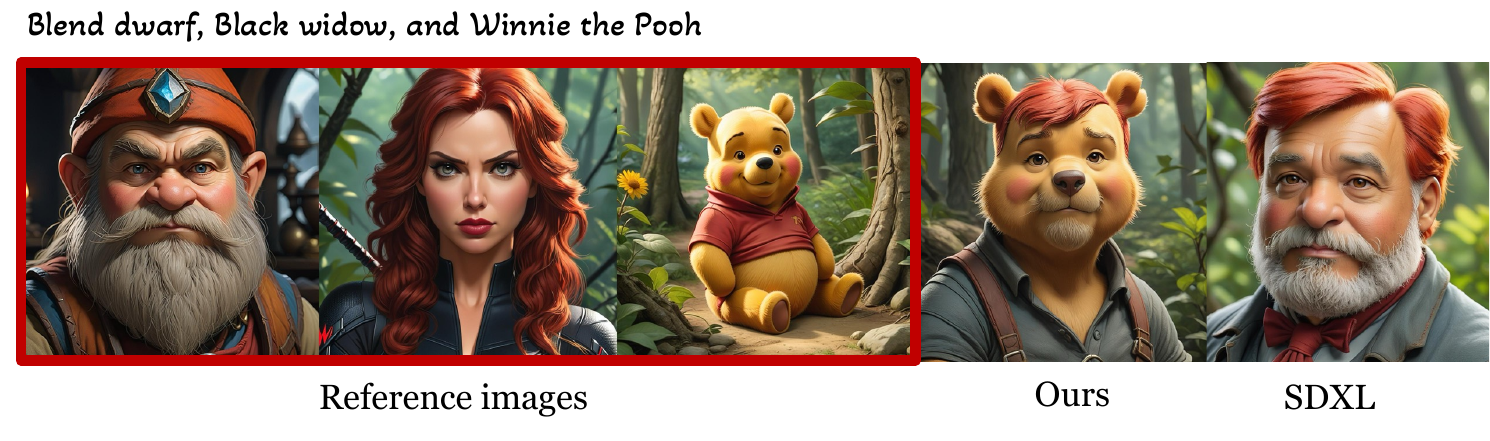}   
    \caption{
    Blending \textbf{three} distinct subjects, we use the same prompt—"a portrait of Winnie the Pooh with red hair and a gray beard"—for both SDXL and \ours. However, SDXL significantly downplays the features of Winnie the Pooh. In contrast, our approach effectively absorbs the features from the reference images, retaining the dwarf's outfit and beard, Black Widow's red hair, and Winnie's facial structure.
    }  \label{fig:blend_3reference}
\end{figure}

 \textbf{Blend multiple images:} We show additional blended images using multiple reference images in \cref{fig:blend2}, 
 \cref{fig:blend3}, and \cref{fig:blend_3reference}. In particular, \cref{fig:blend2}, \cref{fig:blend3} utilize two reference images, and \cref{fig:blend_3reference} blends three reference images.

\begin{figure}[h!]
  \centering
    \makebox[\linewidth]{\includegraphics[width=1\linewidth]{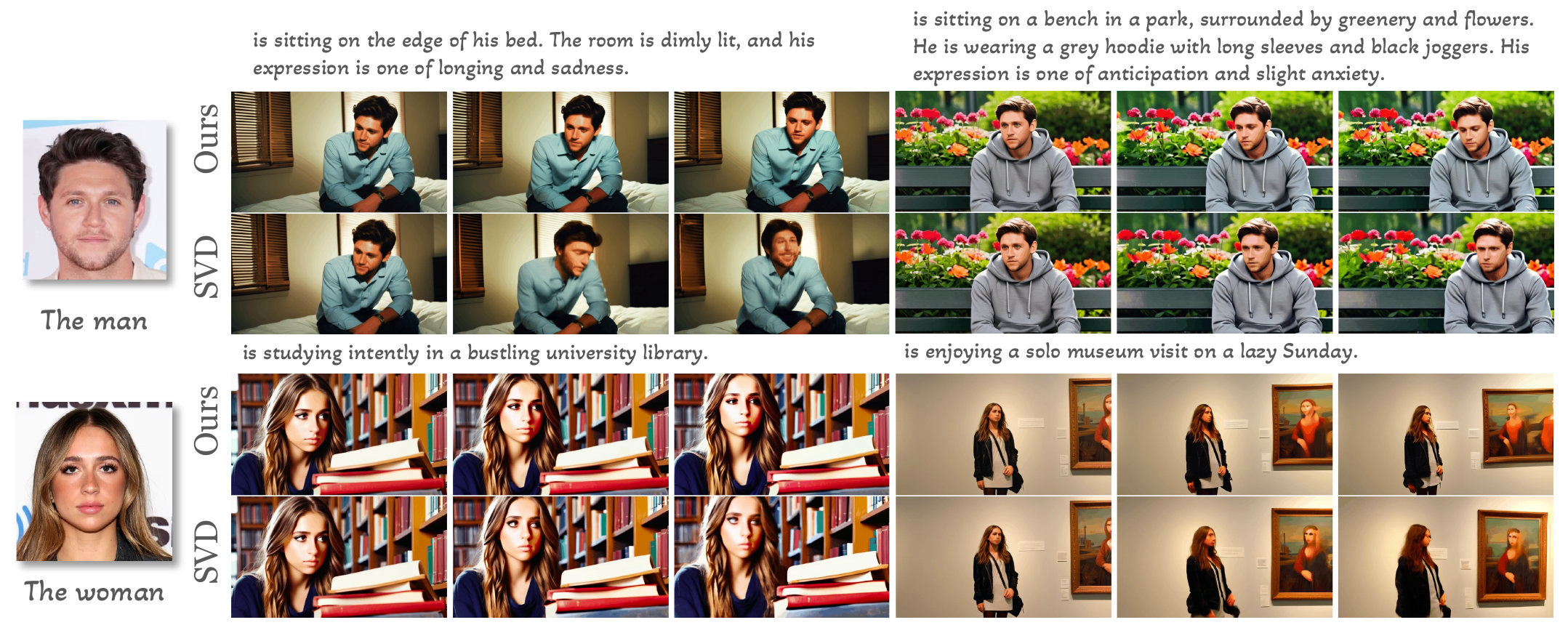}}
  \caption{Additional personalized video comparison. The original videos can be viewed \href{https://refdrop-anonymouspaper-f37a6c745f264e0ff8b994669d71e9ca5f34d07a.gitlab.io/index.html}{here}.
  }
  \label{fig:person_vid2}
\end{figure}

\textbf{Personalized Video Comparisons:}
We show additional comparison in \cref{fig:person_vid2}. Moreover,
we offer more than 20 original videos in $1024 \times 576$ resolution, accessible via this \href{https://refdrop-anonymouspaper-f37a6c745f264e0ff8b994669d71e9ca5f34d07a.gitlab.io/index.html}{anonymous external link}. On the linked page, the left column displays the video generations of SVD, while the right column features the enhanced SVD results by \ours.

\section{Ablation Study}\label{sec:ablate}

We provide some guidelines on the effect of the coefficient:
\begin{itemize}
    \item \textbf{Consistent Image Generation:} More challenging tasks typically require larger coefficients to ensure consistency. For example, generating human figures, which are more complex, requires coefficients between $[0.3, 0.4]$. In contrast, simpler subjects like fluffy toys or cartoon characters may only need a coefficient of $0.2$ to achieve consistent generation.
    \item \textbf{Blend multiple images:} We find that the coefficients for each reference image, typically falling within the range of $[0.2, 0.4]$, perform effectively.
    \item \textbf{Diverse Image Generation:} We recommend using a coefficient of $c=-0.3$. Lower strengths can impair visual quality and may introduce artifacts.
    \item \textbf{Video Consistency:} The coefficient for video consistency requires more nuanced control; A coefficient of $0.2$ generally suffices, and a larger coefficient may make the video totally static. This sufficiency is likely due to the temporal attention component in VDM, which tends to amplify the effects introduced through self-attention.
\end{itemize}

The effect of reference strength on image generation is in \cref{fig:coeff}.

\section{Additional implementation details}\label{sec:imple}

All experiments were conducted on a single NVIDIA A100 GPU  with 80GB of memory. The generation process for a single image using SDXL requires approximately 5 seconds, whereas generating a video using SVD takes about 30 seconds. Additional details on hyper-parameters for both baseline methods and our approach are provided in \cref{tab:param}.

\begin{table}[h!]
    \caption{Base model and hyper-parameters.}
    \label{tab:param}
    \vspace{0.3cm}
    \centering
    \begin{tabular}{@{}c>{\raggedright}p{3cm}cccc@{}}
        \toprule
        & \textbf{Base model} & \textbf{CFG} & \textbf{\begin{tabular}[c]{@{}c@{}}Our reference\\ strength\end{tabular}} & \textbf{\begin{tabular}[c]{@{}c@{}}IP-Adapter\\ scale\end{tabular}} & \textbf{TLPFF} \\ 
        \midrule
        \cref{sec:consist} & Protovision-XL & 5 & 0.3$\sim$0.4 & 0.6 & N/A \\
        \cref{sec:blend} & Protovision-XL & 5 & 0.2$\sim$0.4 & N/A & N/A \\
        \cref{sec:diverse} & Protovision-XL & 5 & -0.3 & -0.6 & N/A \\
        \cref{sec:consist_video} & SVD-img2vid-xt-1-1 & 2.5 & 0.2 & N/A & \begin{tabular}[c]{@{}c@{}}Gaussian filter\\ Stop frequency = 0.5\end{tabular} \\
        \cref{sec:pv} & SVD-img2vid-xt-1-1 & 2.5 & 0.2 & N/A & N/A \\
        \bottomrule
    \end{tabular}
\end{table}

\section{Human evaluation details}\label{sec:user_study}

The Google Forms survey contains 5 sections, encompassing a total of 50 questions. Instructions and examples are detailed in attached screenshots for each section.

\begin{figure}[h!]
  \centering
    \centering
    \includegraphics[width=0.9\linewidth]{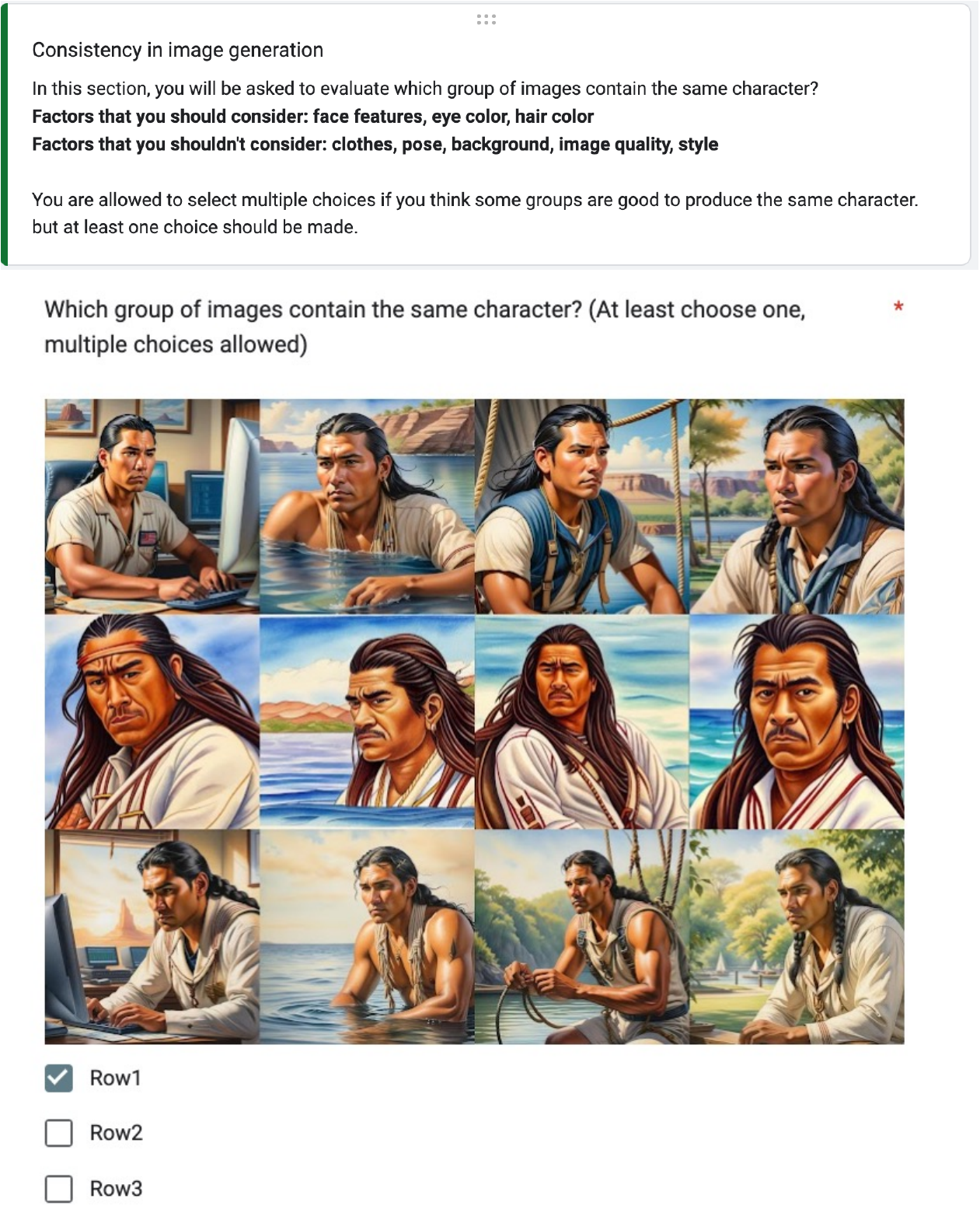}
  \caption{The instruction and example for human evaluation.} \label{fig:h1}
\end{figure}

\begin{figure}[h!]
  \centering
    \centering
    \includegraphics[width=0.8\linewidth]{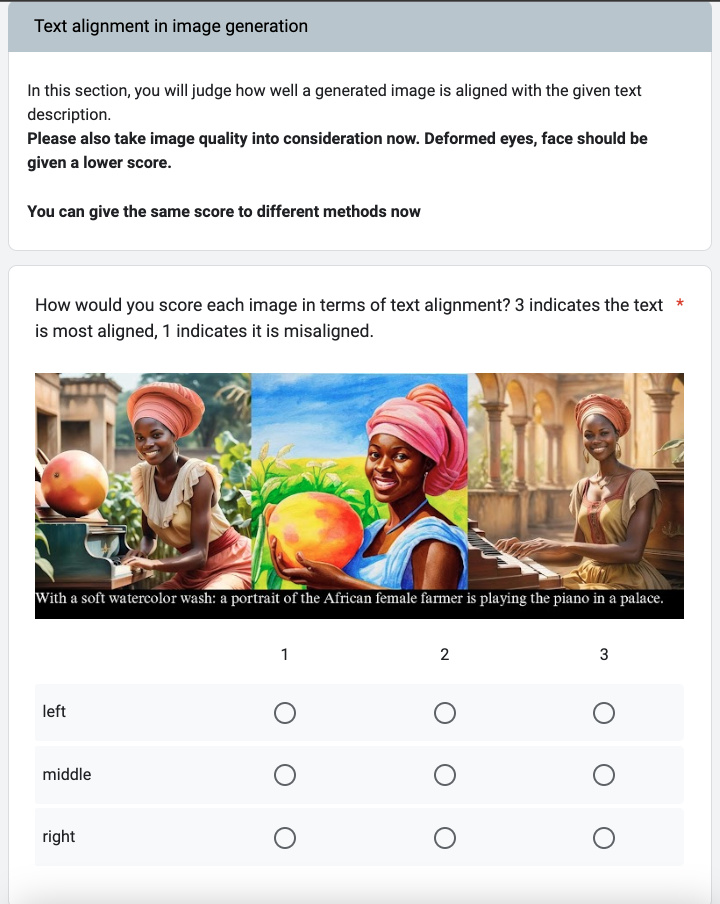}
  \caption{The instruction and example for human evaluation.} \label{fig:h3}
\end{figure}

\begin{figure}[h!]
  \centering
    \centering
    \includegraphics[width=0.8\linewidth]{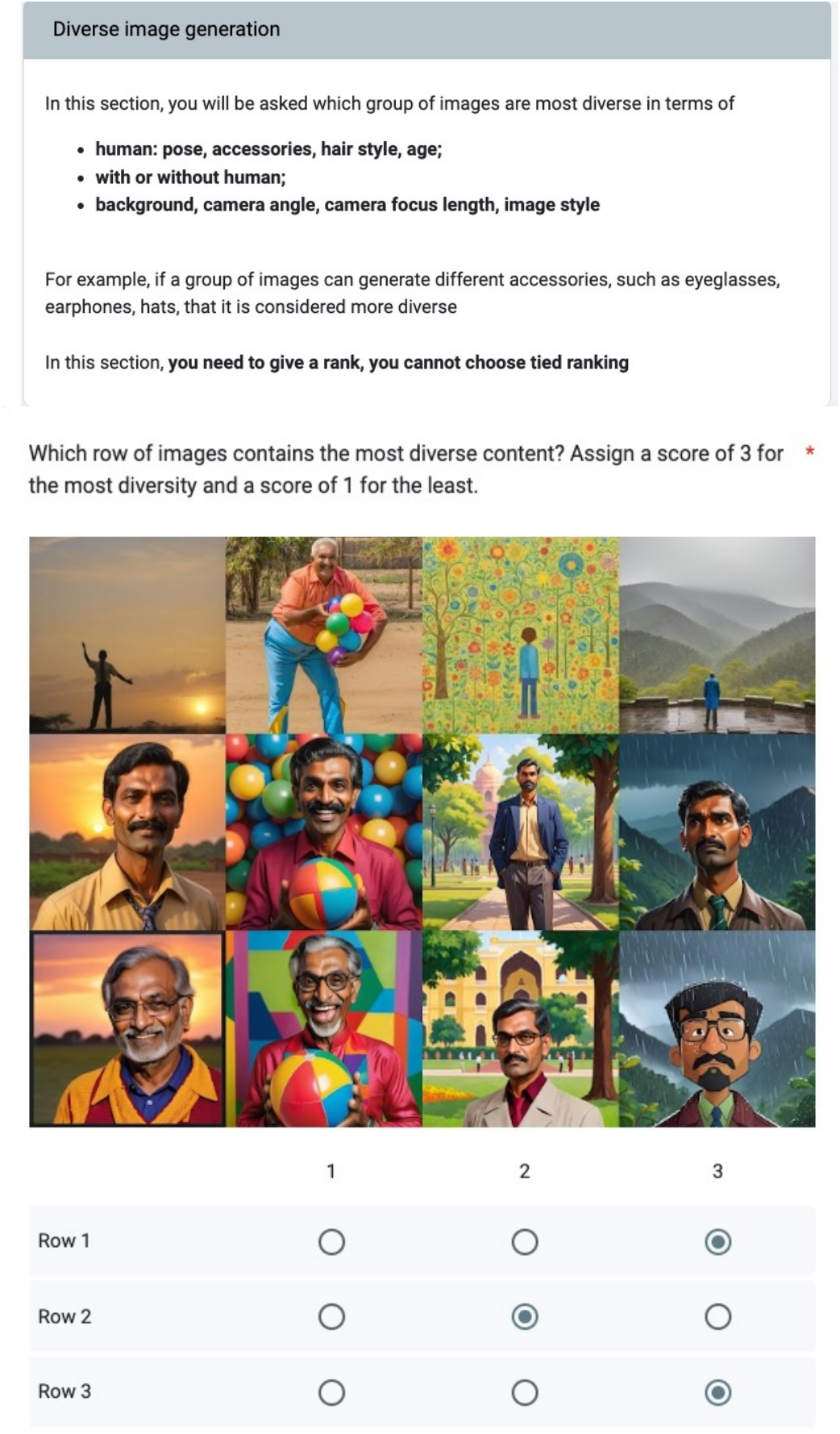}
  \caption{The instruction and example for human evaluation.} \label{fig:h2}
\end{figure}

\begin{figure}[h!]
  \centering
    \centering
    \includegraphics[width=0.8\linewidth]{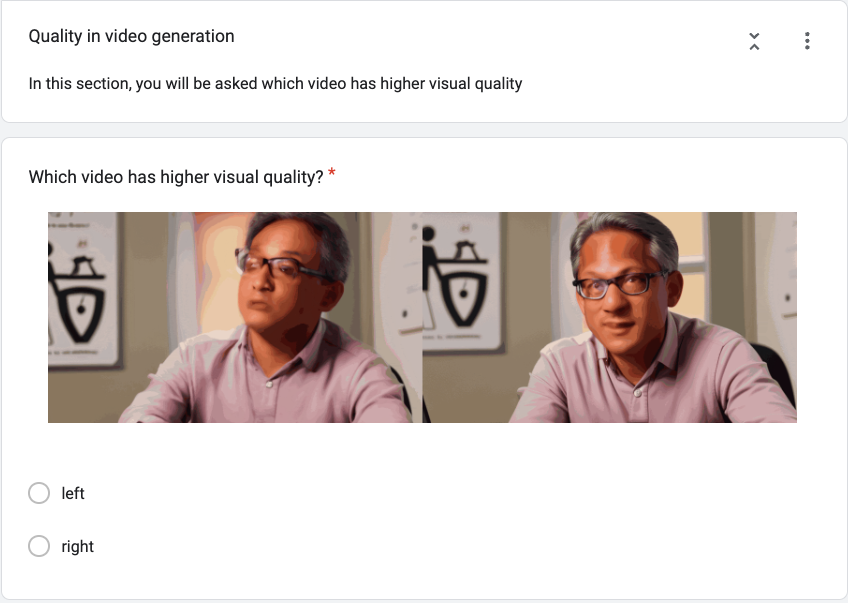}
  \caption{The instruction and example for human evaluation.} \label{fig:h4}
\end{figure}

\begin{enumerate}
    \item \textbf{Visual Consistency in Consistent Image Generation}: Participants evaluate visual consistency across five images of the same subject, for example, ``Native American sailor'' produced by different methods,  Methods that maintain character consistency are scored 1; others receive a score of 0. Detailed instructions are provided in \cref{fig:h1}.
    \item \textbf{Text Alignment in Consistent Image Generation}: Respondents assess the alignment of text with the corresponding image for each method, assigning a score from 1 to 3, where a higher score indicates better alignment. Detailed instructions are provided in \cref{fig:h3}.
    \item \textbf{Visual Diversity in Diverse Image Generation}: Like the first section, participants rate the diversity in five images of the same subject across different methods. They assign a score from 1 to 3, with a higher score indicating greater diversity. Detailed instructions are provided in \cref{fig:h2}.
    \item \textbf{Text Alignment in Diverse Image Generation}: This section mirrors Section 2 but in the context of diverse image generation. Participants rate text-image alignment on a scale from 1 to 3. Detailed instructions are provided in \cref{fig:h3}.
    \item \textbf{Personalized Video Quality}: Participants evaluate the quality of videos generated with the same random seed by different methods. Methods that are chosen for higher quality receive a quality score of 1; others receive a score of 0. Detailed instructions are provided in \cref{fig:h4}.
\end{enumerate}

We aggregated scores from all sections and display the results in \cref{fig:human}. \textbf{In total, each participant provided 140 ratings, resulting in 6,160 ratings from 44 participants.}

\section{Licenses}\label{sec:license}

Pretrained models:
\begin{itemize}
    \item ProtoVision-XL\footnote{\url{https://huggingface.co/stablediffusionapi/protovision-xl-high-fidel}}~\citep{podell2023sdxl} CreativeML Open RAIL++-M License
    \item Stable-Video-Diffusion-img2vid-xt-1-1\footnote{\url{https://huggingface.co/stabilityai/stable-video-diffusion-img2vid-xt-1-1}}~\citep{blattmann2023stable} CreativeML Open RAIL++-M License
    \item BLIP Diffusion\footnote{\url{https://huggingface.co/salesforce/blipdiffusion}}~\citep{li2024blip} Apache 2.0 License   
    \item IP-Adapter-SDXL\footnote{\url{https://huggingface.co/h94/IP-Adapter/blob/main/sdxl_models/ip-adapter_sdxl.bin}}~\citep{ye2023ipa} Apache 2.0 License   
    \item InstantID\footnote{\url{https://huggingface.co/InstantX/InstantID}}~\citep{wang2024instantid} Apache 2.0 License       
\end{itemize}

Codebase:

\begin{itemize}
    \item diffusers 0.25.1 \footnote{\url{https://github.com/huggingface/diffusers}}~\citep{diffusers} Apache 2.0 License
    \item EvalCrafter \footnote{\url{https://github.com/EvalCrafter/EvalCrafter}}~\citep{liu2023evalcrafter} No license found
\end{itemize}

Metric models:

\begin{itemize}
    \item OpenCLIP\footnote{\url{https://huggingface.co/laion/CLIP-ViT-g-14-laion2B-s12B-b42K}}~\citep{Radford2021LearningTV,ilharco_gabriel_2021_5143773} MIT License    
    \item DreamSim\footnote{\url{https://dreamsim-nights.github.io/}}~\citep{fu2023dreamsim} MIT License  
    \item LPIPS 1.0\footnote{\url{https://lightning.ai/docs/torchmetrics/stable/image/learned_perceptual_image_patch_similarity.html}}~\citep{zhang2018unreasonable} BSD-2-Clause license
\end{itemize}

\newpage

